\documentclass{article}

\usepackage{microtype} 
\usepackage{graphicx}
\usepackage{subcaption}
\usepackage{booktabs} 
\usepackage{hyperref}

\usepackage[scaled=0.01]{beramono}
\usepackage{amsfonts}       
\usepackage{nicefrac}       
\usepackage{xcolor}         
\usepackage{multirow}
\usepackage[utf8]{inputenc} 
\usepackage[T1]{fontenc}    
\usepackage{courier} 
\usepackage{algorithm}

\usepackage[preprint]{icml2026}

\usepackage{amsmath}
\usepackage{amssymb}
\usepackage{mathtools}
\usepackage{amsthm}

\usepackage[capitalize,noabbrev]{cleveref}

\theoremstyle{plain}
\newtheorem{theorem}{Theorem}[section]
\newtheorem{proposition}[theorem]{Proposition}
\newtheorem{property}[theorem]{Property}

\theoremstyle{definition}
\newtheorem{definition}[theorem]{Definition}

\theoremstyle{remark}

\usepackage[textsize=tiny]{todonotes}

\def\Vertcohirf{\text{VertCoHiRF}{}}
\def\bbX{\mathbb{X}}
\def\K{\text{K}}
\def\L{\text{L}}
\newcommand{\R}{\mathbb R}
\newcommand{\BCM}{\text{BCM}{}}
\newcommand{\Lconcat}{\mathbf{\mathbb{L}}}

\newcommand{\Xa}{X_a}
\newcommand{\vP}{\text{P}}

\newcommand{\MedoidsScore}{{\text{ChooseMedByScore}}}
\newcommand{\MedoidsSort}{{\text{Clust\&MedSort}}}
\newcommand{\GetClusters}{{\text{GetClusters}}}
\newcommand{\UpdateParents}{{\text{UpdateParents}}}
\newcommand{\MListconcat}{{\text{$\mathbb{M}$edList}}}
\newcommand{\GetFinalLabels}{{\text{GetFinalLabels}}}
\newcommand{\MList}{{\text{MedList}}}

\newcommand{\rank}{{\text{rank}{}}}
\newcommand{\ARI}{{\text{ARI}{}}}
\newcommand{\VPC}{{\text{DPFMPS}{}}}
\newcommand{\VWay}{{\text{DPFMPS-P2Est}{}}}
\newcommand{\DistrKMeans}{{\text{Distr.KMeans}{}}}
\newcommand{\Coreset}{{\text{CoresetKMeans}{}}}

\icmltitlerunning{VertCoHiRF: decentralized vertical clustering beyond k-means}

\begin{document}

\twocolumn[ \icmltitle{VertCoHiRF: decentralized vertical clustering beyond k-means}



  \icmlsetsymbol{equal}{*}

  \begin{icmlauthorlist}
\icmlauthor{Bruno Belucci}{equal,yyy}
\icmlauthor{Karim Lounici}{equal,sch}
\icmlauthor{Vladimir R. Kostic}{equal,xxx}
\icmlauthor{Katia Meziani}{equal,yyy}
  \end{icmlauthorlist}
\icmlaffiliation{yyy}{CEREMADE, University of Paris Dauphine - PSL, Paris, France}  
\icmlaffiliation{sch}{CMAP , Ecole Polytechnique, Palaiseau, FRANCE}
\icmlaffiliation{xxx}{Istituto Italiano di Tecnologia, Italy, France}

\icmlcorrespondingauthor{Bruno Belucci}{bruno.belucci-teixeira@dauphine.eu}
\icmlcorrespondingauthor{Karim Lounici}{karim.lounici@polytechnique.edu}
\icmlcorrespondingauthor{Vladimir R. Kostic}{vladimir.kostic@iit.it}
\icmlcorrespondingauthor{Katia Meziani}{meziani@ceremade.dauphine.fr}
  \icmlkeywords{Machine Learning, ICML, clustering, cohirf, vertical, federated learning, vertcohirf, privacy, collaboration}

  \vskip 0.3in
]



\printAffiliationsAndNotice{}  

\begin{abstract}

Vertical Federated Learning (VFL) enables collaborative analysis across parties holding complementary feature views of the same samples, yet existing approaches are largely restricted to distributed variants of $k$-means, requiring centralized coordination or the exchange of feature-dependent numerical statistics, and exhibiting limited robustness under heterogeneous views or adversarial behavior.
%
We introduce \Vertcohirf, a fully decentralized framework for vertical federated clustering based on structural consensus across heterogeneous views, \textbf{allowing each agent to apply a base clustering method adapted to its local feature space in a peer-to-peer} manner. 
Rather than exchanging feature-dependent statistics or relying on noise injection for privacy, agents cluster their local views independently and reconcile their proposals through identifier-level consensus. Consensus is achieved via decentralized ordinal ranking to select representative medoids, progressively inducing a shared hierarchical clustering across agents.
%
Communication is limited to sample identifiers, cluster labels, and ordinal rankings, providing \textbf{privacy by design} while supporting overlapping feature partitions and heterogeneous local clustering methods, and yielding an interpretable shared \textbf{Cluster Fusion Hierarchy (CFH)}
that captures cross-view agreement at multiple resolutions.
We analyze communication complexity and robustness, and experiments demonstrate competitive clustering performance in vertical federated settings.
\end{abstract}

\section{Introduction}

Federated learning has emerged as a principled paradigm for collaborative data analysis under privacy and governance constraints \cite{albrechtHowGDPRWill2016}, enabling multiple organizations to learn from distributed data without sharing raw data \cite{yeVerticalFederatedLearning2025}. While federated learning has achieved remarkable success in supervised settings, unsupervised problems, and clustering in particular, remain comparatively underexplored, a gap that is especially pronounced in vertical federated learning (VFL), where different parties hold complementary feature subsets describing the same set of individuals.

Vertical federated clustering naturally arises in many real-world applications \cite{liReviewApplicationsFederated2020}. In healthcare, hospitals may hold different clinical measurements for overlapping patient populations and seek to identify disease subtypes. In finance, institutions sharing customer identifiers but maintaining heterogeneous transaction records aim to uncover latent behavioral patterns. In such settings, access to the full feature space would substantially improve clustering quality, yet privacy regulations, contractual constraints, and competitive considerations prohibit the sharing or central aggregation of raw data.

\textbf{Related Work}  Existing approaches to clustering in VFL face a persistent tension between \emph{\textbf{privacy, communication efficiency, scalability, and decentralization}}. Existing methods in the literature are built around adaptations of the $k$-means objective, owing to its computational simplicity and ease of distribution.   Distributed $k$-means approaches \cite{ding16,zhu_f3km_2023} rely on iterative exchange of numerical statistics and typically assume the presence of a central coordinator, leading to strong trust assumptions. Coreset-based methods \cite{Huang2022} follow a different communication strategy by transmitting compact summaries, but remain tightly coupled to the $k$-means objective and require sharing feature-dependent numerical information.\\
Privacy-preserving VFL clustering has been studied through differentially private  frameworks such as \VPC{}, including variants relying on two-way communication protocols \cite{liDifferentiallyPrivateVertical2023}. These methods provide formal privacy guarantees by injecting noise into shared statistics or representations. While effective from a privacy standpoint, the resulting noise introduces an inherent privacy–utility trade-off and requires centralized coordination, iterative optimization, and objective-specific protocols. As a result, existing methods remain limited in their ability to support flexible clustering paradigms beyond $k$-means, or to operate efficiently in fully decentralized settings.


\textbf{Contribution} A key limitation shared by most existing approaches is their reliance on a \textbf{single global clustering objective} and on the exchange of numerical quantities derived from private features. In contrast, real-world data often exhibit heterogeneous and non-convex structures, and different feature subsets may call for different local clustering strategies. Motivated by this observation, we propose a different perspective: rather than distributing a specific clustering objective, vertical federated clustering can be viewed as a problem of \textbf{reaching consensus over locally inferred structure}.\\
Building on this perspective, we introduce \Vertcohirf{}, a decentralized clustering framework for VFL that extends the consensus-based hierarchical structure discovery principles of {CoHiRF} \cite{Cohirf}  to vertically partitioned data. {While \Vertcohirf{} draws inspiration from \textsc{CoHiRF}, it is not a direct extension nor a plug-and-play adaptation. In CoHiRF, consensus is obtained by aggregating multiple randomized partitionings computed over a single feature space. In contrast, \Vertcohirf{} operates in a vertical federated setting where each agent observes a distinct and possibly overlapping subset of features and produces structurally heterogeneous partitions. As a result, the consensus mechanism no longer aims at stabilizing a single clustering objective, but at aggregating independent structural signals arising from heterogeneous views through identifier-level communication.} Unlike existing VFL clustering methods, \Vertcohirf{} does not distribute or approximate a global objective such as $k$-means. Instead, each agent independently applies a clustering method of its choice to its local feature view, and the global structure emerges through an iterative consensus mechanism operating solely on \textbf{identifier-level information}.

At each iteration, agents exchange only cluster assignments and ordinal rankings of sample identifiers, without sharing features, distances, centroids, embeddings, or any intermediate numerical statistics. 
%
Only sample groupings that are not contradicted by any heterogeneous view are retained, while view-specific artifacts are eliminated. In this sense, \Vertcohirf{} implements a veto-based consensus mechanism.
Representative medoids are selected through decentralized ranking aggregation, progressively reducing the problem size and inducing an interpretable hierarchical structure.

%
This design yields several desirable properties. Privacy preservation is inherent to the protocol, as feature values are never transmitted. The framework is fully decentralized and operates in a peer-to-peer fashion without trusted coordinators. Communication costs decrease rapidly across iterations as the number of active samples drops. Finally, by decoupling collaboration from any single clustering objective, \Vertcohirf{} naturally supports heterogeneous, structure-aware clustering and exhibits robustness to unreliable or adversarial agents.

Our contributions are summarized as follows:

\textbf{(i) Structural consensus for VFL clustering.}
We reformulate clustering in VFL as a problem of reaching consensus over locally inferred cluster structures, rather than sharing features or optimizing a single global objective.

\textbf{(ii) A fully decentralized framework.}
We introduce \Vertcohirf{}, a fully decentralized peer-to-peer framework for VFL clustering that operates exclusively through identifier-level communication, without relying on central coordinators, trusted servers, or the exchange of feature-dependent numerical statistics.

\textbf{(iii) Flexible and structure-aware collaboration.}
\Vertcohirf{} allows agents to employ base clustering methods adapted to their local feature modalities, enabling the recovery of complex cluster structures beyond $k$-means–based VFL approaches.

\textbf{(iv) Communication-efficient and privacy-preserving design.}
We show that \Vertcohirf{}’s identifier based communication achieves linear complexity, while preserving privacy by design without noise injection or cryptographic primitives.

\textbf{(v) Interpretability.}
The proposed framework naturally produces an explicit and shared \emph{Cluster Fusion Hierarchy} (CFH), providing an interpretable hierarchical organization of the consensus clustering.

\textbf{(vi) Robustness analysis.}
The proposed protocol is robust to Byzantine agents by design, as it relies on an agent-level veto mechanism rather than a majority-based consensus.

\textbf{(vii) Empirical evaluation.}
Experiments on synthetic and real-world datasets demonstrate that \Vertcohirf{} achieves competitive clustering quality in VFL settings, and highlights the benefits of structure-adapted local clustering methods.

\section{The \Vertcohirf{} framework}


Consider $n$ samples shared across $A$ agents in a vertical federated setting. Each agent $a \in [A]$ observes the complete set of samples through a partial feature view: agent $a$ accesses a feature subset $\mathcal{P}_a \subseteq \{1,\ldots,p\}$, yielding a local data matrix $\bbX_a \in \mathbb{R}^{n \times p_a}$. The global feature matrix $\bbX \in \mathbb{R}^{n \times p}$ remains distributed, and no agent has access to the full feature space.

We allow overlapping feature subsets, meaning that $\mathcal{P}_a \cap \mathcal{P}_b$ may be non-empty. This captures practical scenarios where certain attributes are shared across institutions, and is more general than standard VFL formulations that assume disjoint feature partitions.

The objective is to discover the clustering structure induced by the complete feature space through collaborative learning, while ensuring that no agent gains access to features outside its local subset $\mathcal{P}_a$. Conventional approaches typically exchange data-dependent quantities such as intermediate representations, centroids, distance matrices, or embeddings, which can create privacy and security vulnerabilities. \Vertcohirf{} instead enforces consensus on cluster structure using identifier-level communication.

\subsection{Core idea: Structural consensus}

\Vertcohirf{} is built on a simple principle: if a cluster reflects genuine structure in the full feature space, then it should be consistently observable across heterogeneous and partial feature views. Conversely, groupings induced by view-specific noise or artifacts are unlikely to be reproduced across agents observing different subsets of features.\\
This principle is particularly suited to VFL, where all agents observe the same samples through distinct feature projections. Although no agent has access to the complete feature space, stable global structure should manifest as recurrent patterns across independent local clusterings.\\
Crucially, \Vertcohirf{} does not aim to enforce a shared global geometry or a single clustering objective across agents. Instead, collaboration is achieved through \textbf{structural consensus}: agents are free to use heterogeneous local clustering methods, and agreement is enforced only at the level of discrete cluster structure.
%


\Vertcohirf{} implements this idea by shifting collaboration from sharing numerical quantities to identifying compatible structural relations through identifier-level communication. Each agent independently performs clustering using its own local feature view and communicates only cluster labels and ordinal rankings. Global structure emerges by retaining only groupings that are not contradicted across agents, without requiring the local clusterings to be identical.

\subsection{Consensus Protocol}

\Vertcohirf{} operates through an iterative, fully decentralized protocol. At each iteration $e$, agents alternate between local structure discovery and consensus phases based on identifier-level exchanges, progressively refining the clustering while reducing the number of active samples.

Initially, all samples are considered active medoids. At iteration $e$, only the active representatives from previous iteration $e-1$ participate in local clustering. Non-selected samples are permanently attached to their most recent representative, which induces a hierarchical organization of the data. \\
At iteration $e$, $\K^{(e-1)}$  denote the set of identifiers of the $n^{(e-1)}$ active  medoids.  The protocol then proceeds in two communication phases based on identifier-level exchanges (see Algorithm~\ref{alg:vertcohirf}).

\textbf{Phase~1 }\textit{(Structural consensus).}
First, each agent $a$ applies a local Base Clustering Method (BCM) to the active medoids using only its local observations $\bbX_a^{(e)} = \bbX_a[\K^{(e-1)}, :\,]$. This corresponds to a single local CoHiRF-style step, rather than a full local hierarchical loop. Each agent then \textbf{broadcasts} its resulting cluster labels $\L_a^{(e)}$, indexed by shared sample identifiers. The labels produced by all agents are concatenated to form a joint structural code $\L^{(e)}$, with one component per agent. Two samples are assigned to the same cluster if and only if they share the same code. This construction allows each agent to contribute heterogeneous structural information, enabling multimodal clustering without requiring consistency between
local partitions.

\textbf{Phase~2 }\textit{(Representative consensus).}
For each consensual cluster, agents rank its samples according to their local feature views and \textbf{exchange} only ordered identifier lists. These rankings are then aggregated through a score-based rule to select a single representative per cluster. For each cluster $\mathcal{C}_k^{(e)}$, a medoid candidate $m$ is ranked by each agent $a$, and its ranking position $\rank_a(m)$ contributes to the global score 
$
S(m) = \sum_{a=1}^A \rank_a(m).
$ 
Medoids minimizing this score become the cluster representatives and form the active set for the next iteration.

\subsection{Algorithm description}

{All local clustering and hierarchy update procedures used in Algorithm~\ref{alg:vertcohirf} are inherited from CoHiRF \cite{Cohirf}, except for the representative selection mechanism introduced in Phase~2, which is specific to \Vertcohirf{}.}

Algorithm~\ref{alg:vertcohirf} provides a pseudo-code description of the protocol executed by each agent.

\begin{algorithm}[h]
\caption{[\Vertcohirf\,for agent $a$]}
\label{alg:vertcohirf}
\begin{algorithmic}
\REQUIRE $X_a \in \R^{n \times p_a},\, \BCM_a,\, R_a,\, q_a,\, \textrm{max\_iter}$
\ENSURE Hierarchical structure $\{\Xa^{(e)}\}_{e=0}^{E}$, (CFH)
\STATE $\vP,\, \K^{(0)}\gets (1:n),\,\,(1:n)$
\STATE $e, n^{(0)} \gets 0, n$
\WHILE  {$(n^{(e-1)} \neq n^{(e)})$  \textbf{and} $(e < \text{max\_iter})$}
    \STATE $X_a^{(e)} \gets X_a[\K^{(e-1)}, \,:\,]$
    \STATE $\L_a^{(e)} \gets \GetClusters(X_a^{(e)}, \BCM_a, R_a, q_a)$
    \STATE $\textcolor{blue}{\textbf{Broadcast}}\{ \L_a^{(e)}\}$);\,\,\textbf{\textcolor{blue}{ Collect}}$\{\L_{a'}^{(e)}\}_{a'\neq a}$    
    \STATE $\Lconcat^{(e)} \gets \textbf{Concat}\{\L_{a'}^{(e)}\}_{a'\neq a}$
    \STATE $\MList_a^{(e)},\, \L^{(e)} \gets \MedoidsSort(X_a^{(e)}, \Lconcat^{(e)})$
    \STATE $\textcolor{blue}{\textbf{Broadcast}}\{\MList_a^{(e)}\}$);\,\,\textbf{\textcolor{blue}{ Collect}}$\{\MList_{a'}^{(e)}\}_{a'\neq a}$
    \STATE $\MListconcat^{(e)} \gets \textbf{Store}\{\MList_{a}^{(e)}\}_{ a}$
    \STATE $\K^{(e)}\gets\MedoidsScore(\MListconcat^{(e)})$    
    \STATE $n^{(e)} \gets |\K^{(e)}|$
    \STATE $\vP \gets \UpdateParents(\vP, \L^{(e)}, \K^{(e-1)}, \K^{(e)})$
    \STATE $\K^{(e-1)},\, n^{(e-1)},\, e \gets \K^{(e)},\, n^{(e)}, \,e+1$
\ENDWHILE 
\STATE $\L \gets \GetFinalLabels (\vP, \K^{(E)})$
\STATE \textbf{Return} $\{\Xa^{(e)}\}_{e=0}^{E}$, CFH constructed from $\vP$
\end{algorithmic}
\end{algorithm}

At each iteration, the selected medoids form the active set $\K^{(e)}$ for the next iteration. Consensus clusters are summarized by representative medoids, progressively reducing the effective problem size. The protocol iterates until the active set stabilizes, that is, until $\K^{(e)} = \K^{(e-1)}$, or until a predefined maximum number of iterations is reached.

All non-selected samples are permanently assigned as inactive children of their corresponding medoid.  These assignments are determined by the consensual cluster labels obtained at iteration $e$ and are tracked through parent pointers. The resulting parent-child relationships define a hierarchical clustering structure. Upon convergence, this structure can be directly interpreted as a \textbf{Cluster Fusion Hierarchy (CFH)}, which records all successive cluster fusions induced by the algorithm.

Rather than producing a single flat partition, and unlike traditional dendrograms,
the CFH provides a multi-resolution representation of cluster relationships without requiring the number of clusters to be specified in advance. By iterating this process, \Vertcohirf{} builds a bottom-up hierarchy that progressively transforms locally consistent label relations into a global, interpretable structure.

\section{Theoretical analysis}
\label{sec:theory}

\subsection{Communication complexity}
\label{sec:comm}

At each iteration, each agent applies a local clustering method to the active
sample set $\K^{(e-1)}$. \Vertcohirf{} does not introduce additional local computation beyond a single local clustering step. Communication is restricted to the exchange of cluster labels and ranked lists of sample identifiers. Since the active set shrinks monotonically, the total communication cost is dominated by the first few iterations.

We analyze the communication cost of \Vertcohirf{} in a fully decentralized peer-to-peer setting.
\begin{proposition}[Communication complexity]
\label{prop:comm}
At iteration $e$, the communication cost $b^{(e)}$ of \Vertcohirf{} with $A$ agents expressed in bits
is bounded by
 \begin{eqnarray*}
b^{(e)}&\leq & A(A-1)\left[
n^{(e-1)} \lceil \log_2(\max_a C_a) \rceil\right.\\
&&+\left.
n^{(e)} N_s \lceil \log_2(n) \rceil
\right].
\end{eqnarray*}
\end{proposition}

Here, $n^{(e)}$ denotes the number of active medoids at iteration $e$, $\max_a C_a$ the maximum number of clusters produced locally by any agent, and $N_s$ an upper bound on the number of candidate medoids per cluster considered in the ranking phase.
The bound follows from the fact that, during the cluster-consensus and medoid-consensus phases, agents exchange only cluster labels and ranked lists of global sample identifiers. A detailed derivation is provided in Appendix~\ref{app:comm}.
The quadratic communication cost in the number of agents arises from the fully connected communication pattern used in our implementation. Exploring alternative communication topologies is a natural extension, but is beyond the scope of this work.

Overall, \Vertcohirf{} achieves communication complexity that scales linearly with the number of samples, up to logarithmic factors, and quadratically with the number of agents. Since communication relies exclusively on identifier-level messages and does not require a central coordinator, \Vertcohirf{} is well suited for large-scale decentralized vertical federated clustering.

\textbf{Convergence and local computation.}
\Vertcohirf{} iterates until the set of active medoids stabilizes. As in CoHiRF, each iteration reduces the number of active representatives, inducing a hierarchical clustering structure through parent-child relations. From a computational standpoint, \Vertcohirf{} does not increase the local algorithmic cost. At each iteration, each agent performs a single CoHiRF-style local step on its own feature subset, prior to inter-agent coordination in Phase~1. No full local hierarchical loop is executed. We refer to~\cite{Cohirf} for a detailed analysis of the per-iteration computational complexity.\\
Overall, \Vertcohirf{} achieves communication complexity that is linear in the number of samples up to logarithmic factors, while remaining fully decentralized and relying exclusively on identifier-level messages.

\subsection{Identifier-level privacy guarantees}
\label{sec:privacy}

A central design principle of \Vertcohirf{} is to prevent feature leakage by construction rather than through noise-based protection. All communication between agents is strictly restricted to sample identifiers, cluster labels, and ordinal rankings. In particular, no numerical quantities depending on feature values, such as distances, centroids, gradients, or embeddings, are ever exchanged.

\textit{Our privacy notion differs from classical Differential Privacy (DP)}. Whereas DP protects sensitive information by adding randomness to feature-dependent outputs, \Vertcohirf{} avoids the transmission of such quantities altogether. Privacy preservation is therefore an inherent property of the protocol, rather than an add-on mechanism. We focus on protecting the feature values held by each agent, rather than hiding sample identities or the resulting cluster structure. Cluster memberships and medoid rankings are assumed to be observable by design, while feature values remain strictly local.

A VFL clustering protocol satisfies \textbf{identifier-level structural privacy} if, for any agent $a$ and any other agent $b \neq a$, the messages received by $a$ depend on the feature matrix $\bbX_b$ only through identifier-level structural information, namely cluster memberships and ordinal rankings.

\begin{definition}[Identifier-level structural privacy]
Let $M_a^{(1:E)}$ denote the complete message history received by agent $a$ over $E$ iterations. For any agent $b \neq a$ and for any two feature matrices $\bbX_b$ and $\bbX_b'$ that induce identical cluster assignments and identical ordinal rankings at all iterations, the induced message distributions are identical:
$$
\mathbb{P}\left(M_a^{(1:E)} \,\middle|\, \bbX_b\right)=
\mathbb{P}\left(M_a^{(1:E)} \,\middle|\, \bbX_b'\right).
$$
\end{definition}

\begin{proposition}[Privacy by construction]
\Vertcohirf{} satisfies identifier-level structural privacy.
\end{proposition}

A formal proof is provided in Appendix~\ref{app:privacy}.

\subsection{Structural robustness induced by strict consensus}
\label{sec:theory_struct}

\Vertcohirf{} relies, during the cluster formation phase~1, on a strict consensus rule: a set of points is retained as a consensus cluster if and only if all agents assign identical labels to the corresponding samples. This conservative design induces a strong structural guarantee in the presence of faulty or Byzantine agents.

\begin{property}[Structural integrity under strict consensus]
\label{prop:cluster_integrity}
Under the strict consensus rule used in Phase~1, no Byzantine agent, nor any coalition of agents, can induce the creation of a consensus cluster that is incompatible with the structure induced by the partitions produced by all honest agents. In particular, enforcing a merge between two points is impossible as soon as at least one honest agent distinguishes them. Byzantine agents can at most prevent or delay legitimate merges, but cannot introduce spurious groupings.
\end{property}

A proof is provided in Appendix~\ref{app:theory_struct}. Importantly, this consensus mechanism operates as a veto-based filter rather than a voting rule. Agents are not required to positively confirm a grouping, it suffices that no agent contradicts it. This design naturally accommodates heterogeneous and partial views, where some agents may lack the information needed to support or reject a given grouping.

Property~\ref{prop:cluster_integrity} has direct implications for the second phase of the protocol. Since Phase~2 operates exclusively on clusters that have already been filtered by strict consensus, all medoid candidates originate from sets that are structurally coherent with respect to the honest agents. In particular, no candidate considered at this stage can correspond to a spurious or structurally inconsistent grouping. The size of consensus clusters is maximal at the first iteration, before any contraction by representative selection occurs. At this stage, clusters are still mostly composed of original data points and directly reflect the structure induced by strict consensus, rather than contamination by erroneous elements.\\
%
As a consequence, Phase~2 does not arbitrate between valid and invalid structures, but performs a \textbf{fine-grained selection} among candidates that are all structurally admissible. In this setting, adversarial behavior cannot introduce new erroneous structures and can at most influence the choice among already coherent representatives, with limited structural impact.\\
As the algorithm proceeds, the progressive contraction induced by medoid selection reduces both the number of active samples and the adversarial surface, further stabilizing the protocol across iterations.


\section{Experiments}
\label{sec:Experiments}

The code for \Vertcohirf{} and to replicate all experiments is available at \hyperlink{https://github.com/BrunoBelucci/vertcohirf}{https://github.com/BrunoBelucci/vertcohirf}. All experimental results are reported according to the following protocol.
For each triplet (dataset, method, metric), we perform an independent hyperparameter optimization targeting the corresponding metric. This ensures that each method is evaluated under conditions that are appropriate for the criterion being reported, rather than under hyperparameters optimized for a different objective. In particular, internal validation metrics are not evaluated using hyperparameters optimized for external agreement, and vice versa, which prevents mixing algorithmic behavior with suboptimal or mismatched hyperparameter settings.

Centralized or server-based $k$-means baselines such as \DistrKMeans{} and \Coreset{} are evaluated in an idealized setting, without differential privacy or cryptographic constraints, 
reflecting their best achievable performance.
In contrast, \VPC{} and \VWay{} rely on differentially private mechanisms based on noise injection or representation sharing, which inherently degrades the clustering signal.

This evaluation choice is conservative with respect to \Vertcohirf{}, whose robustness and privacy properties are intrinsic to the protocol. \Vertcohirf{} does not require prior knowledge of the number of clusters and infers it implicitly through its hierarchical consensus mechanism.

\begin{figure}[ht]
\vskip 0.1in
\begin{center}
\centerline{\includegraphics[width=\columnwidth]{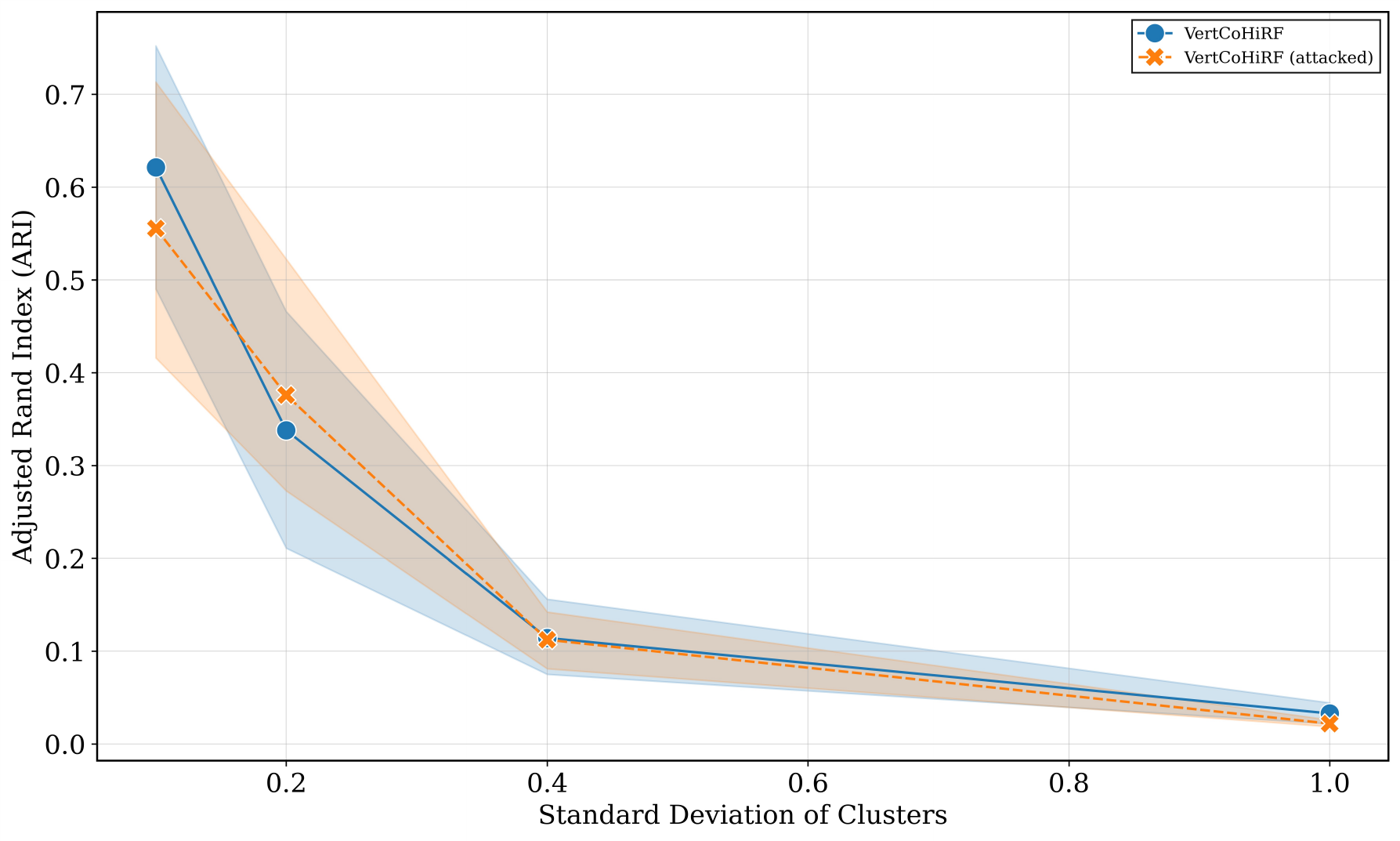}}
\caption{
Empirical robustness of Phase~2 under a single Byzantine attacker. Performance degrades smoothly as the consensus quality decreases with noise $\sigma$.
}
\label{fig:attack}
\end{center}
\end{figure}

We design a controlled synthetic experiment in Figure~\ref{fig:attack} to empirically validate the robustness guarantees established for the medoid selection phase (Phase 2) of \Vertcohirf{} under Byzantine behavior. More details in Appendix~\ref{app:attack}.

\vskip -0.2in

\subsection{A multi-modal clustering setting beyond k-means}
\label{subsec:illustrative_example}

To illustrate a fundamental limitation of VFL clustering methods based on a single
$k$-means objective, we construct a synthetic setting that mirrors realistic collaborative scenarios in which different agents observe heterogeneous data views. In such contexts, the relevant latent structure typically emerges from the interaction between these heterogeneous views rather than from any individual representation.

We generate a single dataset consisting of 1200 observations, where each observation is described by \textbf{two independent feature modalities}. The dataset is designed so that each modality induces a valid but partial clustering structure, while the global clustering structure, defined by their joint interaction, cannot be recovered without \textbf{structural coordination across heterogeneous views}.

(i) The first modality, corresponding to features 1–3, provides a continuous geometric representation in three dimensions. Observations are sampled from the surfaces of two concentric spheres with mean radii equal to 3 and 7, perturbed by radial Gaussian noise with standard deviation 0.3. This modality naturally induces two non-convex clusters and intrinsically incompatible with the assumptions underlying $k$-means, but well suited to geometric or density-based clustering methods such as DBSCAN. (ii) The second modality, corresponding to features 4–5, provides a discrete and structured representation in two dimensions. Each observation is independently associated with a vertex of a square, yielding three discrete clusters. This modality is completely decoupled from the spherical geometry and does not encode any information about it. In contrast to the first modality, its structure is perfectly compatible with $k$-means.
\begin{figure}[h]
\vskip -0.1in
\begin{center}
\centerline{
\begin{subfigure}[b]{0.45\columnwidth}
\centering
\includegraphics[width=\columnwidth]{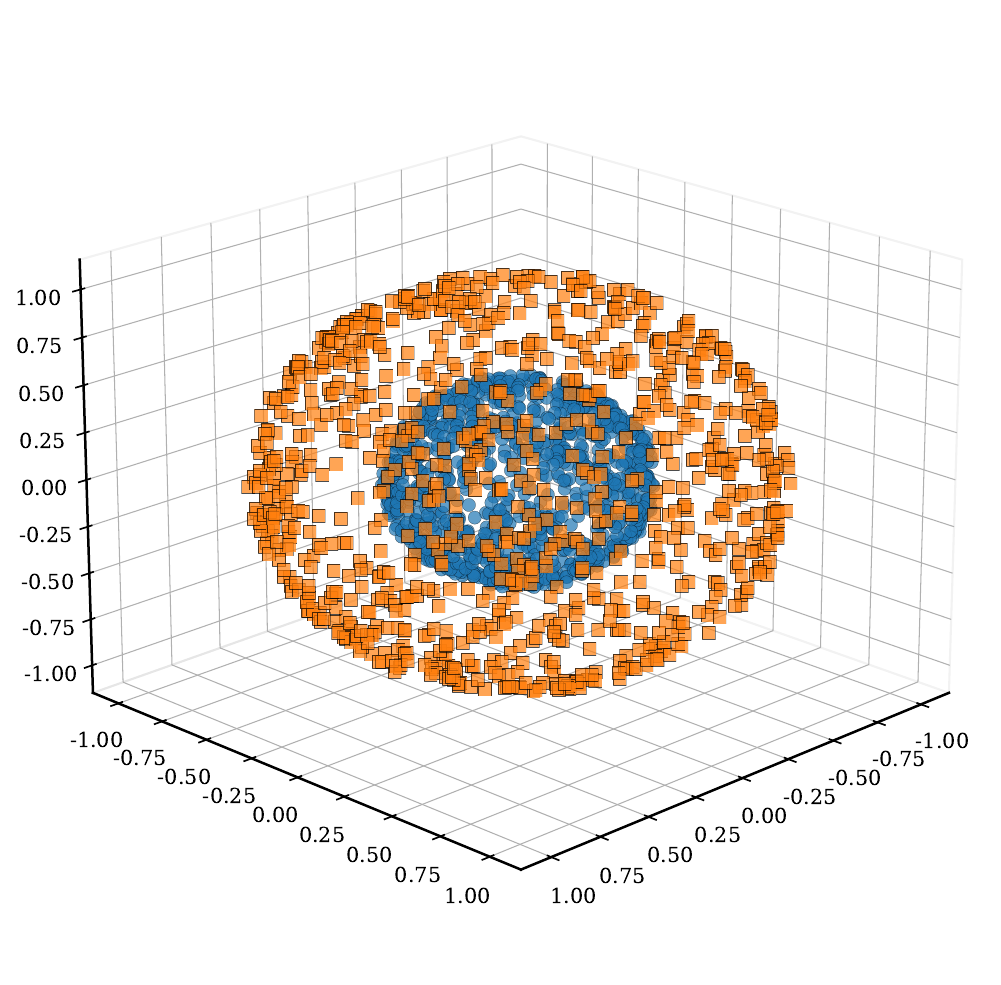}
\caption{Agent$_0$ view (features 1-3): 2 concentric spherical clusters.}
\label{fig:firstmodality}
\end{subfigure}
\hfill
\begin{subfigure}[b]{0.45\columnwidth}
\centering
\includegraphics[width=\columnwidth]{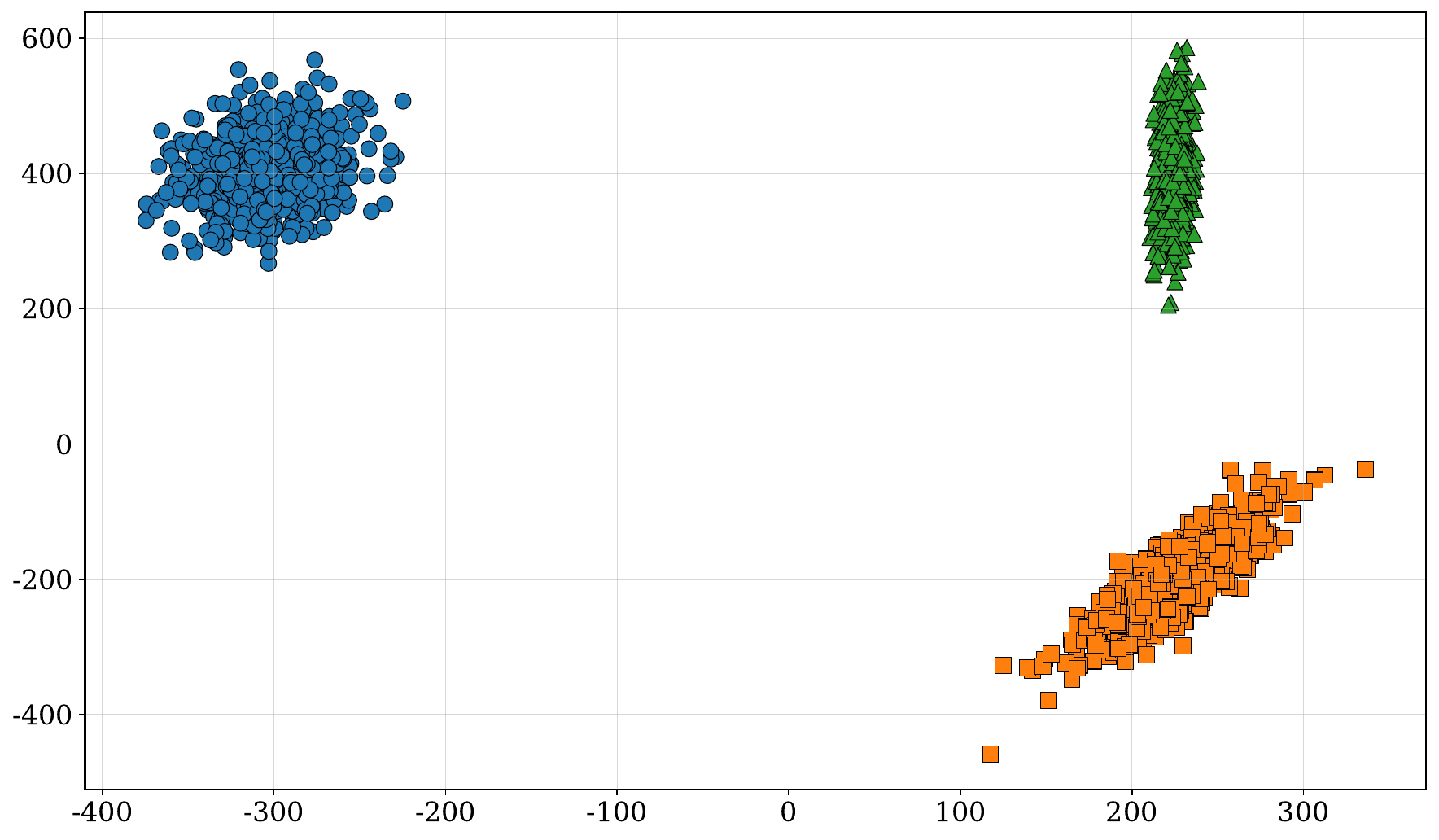}
\caption{Agent$_1$ view (features 4-5): 3 discrete square-based clusters.}
\label{fig:secondmodality}
\end{subfigure}
}
\caption{
Local views of the synthetic multimodal dataset.
Each agent observes a coherent but partial clustering structure in its own feature space.
}
\label{fig:modalities}
\end{center}
\end{figure}
 \vskip -0.1in

Although the two modalities are generated independently, they are jointly observed for each data point. Each observation therefore simultaneously possesses a geometric position on a sphere and an assignment to a square-based structure. Ground-truth labels are defined jointly by the two modalities: an observation belongs to one of the two spheres and to one of the three square-based clusters, and the global label corresponds to the combination of these two assignments. Consequently, the dataset contains six global clusters.

This dataset is multi-modal in a strong sense rather than a simple multi-view setting. Neither modality alone is sufficient to recover the six global clusters: the spherical modality reveals only two groups, while the square modality reveals only three. Each modality captures a clustering structure that is internally consistent but incomplete, and the global structure is accessible only through their combination.
\begin{figure}[h]
\vskip -0.3in
\begin{center}
\centerline{
\begin{subfigure}[b]{0.45\columnwidth}
\centering
\includegraphics[width=\columnwidth]{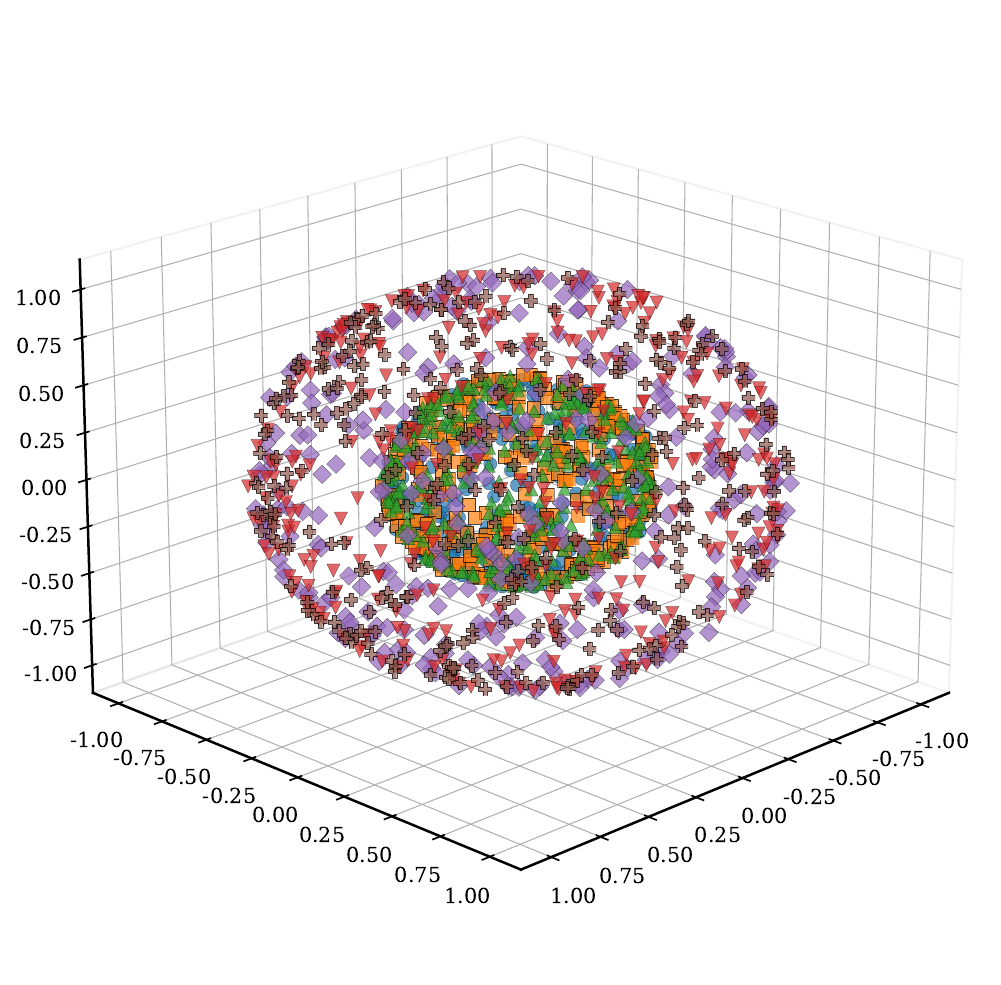}
\caption{Agent$_0$ view : 6 true clusters, 3D view.}
\label{fig:firstmodality6}
\end{subfigure}
\hfill
\begin{subfigure}[b]{0.45\columnwidth}
\centering
\includegraphics[width=\columnwidth]{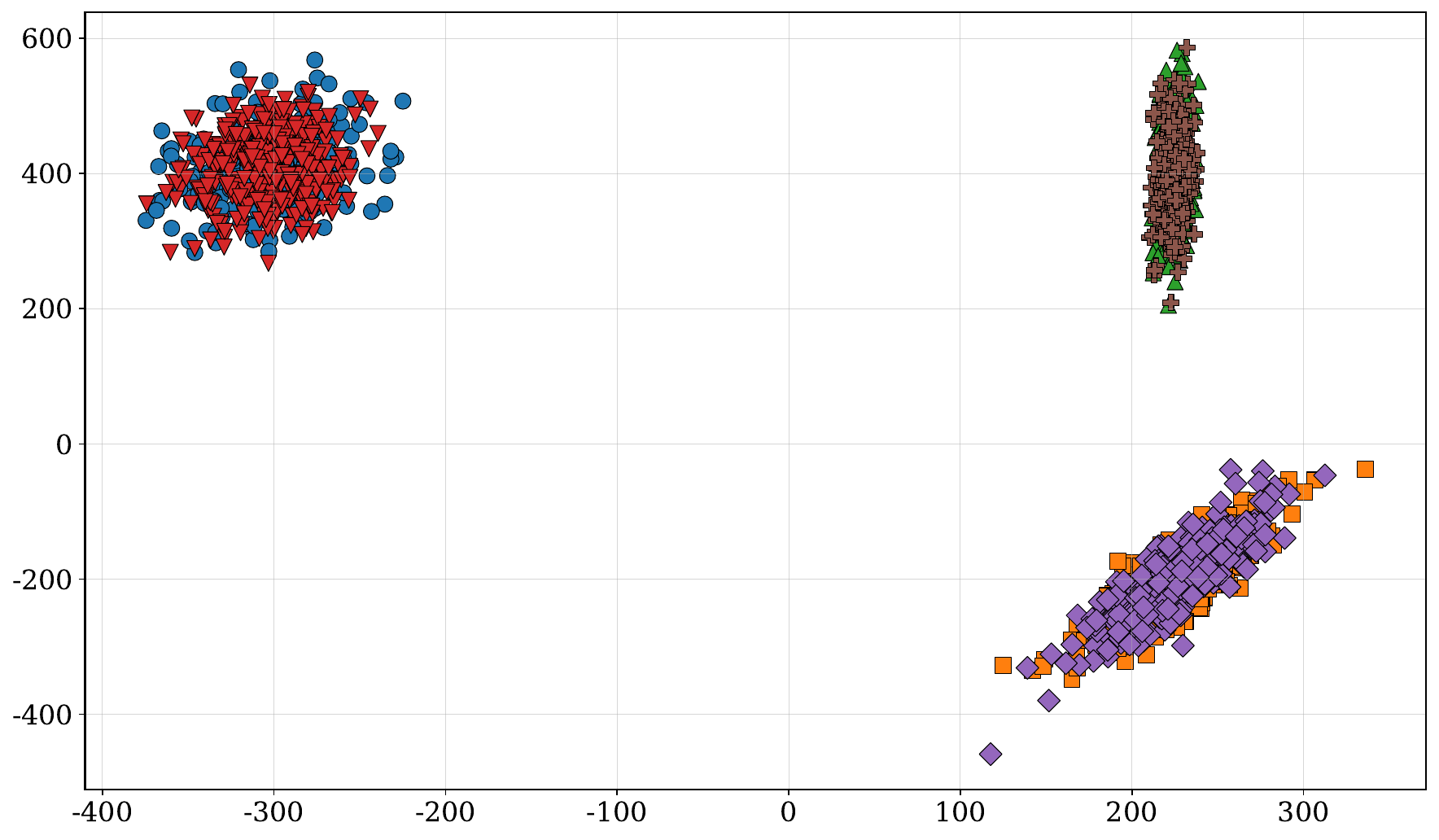}
\caption{Agent$_1$ view : 6 true clusters, 2D view.}
\label{fig:secondmodality6}
\end{subfigure}
}
\caption{
Visualization of the 6-cluster ground-truth partition projected onto each
agent’s feature space.
}
\label{fig:modalities6}
\end{center}
\end{figure}
\vskip -0.3in

From a methodological perspective, $k$-means fails for two distinct reasons. On the spherical modality, its geometric assumptions are violated due to non-convex cluster shapes. On the combined feature space, no Euclidean metric can simultaneously respect the heterogeneous structures induced by the two modalities. Enforcing a single global clustering objective is therefore structurally inappropriate.

This experiment shows that the ground-truth clustering structure may arise from the interaction of multiple independent factors rather than by a single latent criterion. Different modalities may require different \BCM{}, and the global structure does not correspond to any individual view nor by feature concatenation followed by $k$-means. In such settings, \textbf{structural consensus}, rather than objective sharing, is the appropriate primitive for collaborative clustering.


All results are evaluated using \ARI{} with respect to the global multimodal ground-truth partition. The \textcolor{blue}{K-Means} baseline corresponds to the absence of collaboration. The Mean column reports the average \ARI{} across agents; for methods that reach a consensus partition (\DistrKMeans{} and \Vertcohirf), all agents obtain identical scores. Table~\ref{sample-table} reports the performance achieved by each agent after collaboration and reflects both clustering accuracy and inter-agent alignment.
\begin{table}[h]
\caption{
ARI score on the synthetic multimodal dataset.
}
\label{sample-table}
\begin{center}
\begin{small}
\begin{sc}
\begin{tabular}{lcccr}
\toprule
Method & Agent$_0$ & Agent$_1$ & Mean \\
\textcolor{blue}{K-Means} & \textcolor{blue}{0.075216}  &  \textcolor{blue}{0.275385} & \textcolor{blue}{0.175301}\\
\midrule
\DistrKMeans& \underline{0.430624} & 0.430624& \underline{0.430624} \\
\Coreset  & 0.000911 &  0.468107 & 0.234509 \\
\VPC      & 0.114054 &  0.555552 & 0.334803  \\
\VWay     & 0.056390  & \underline{0.565406} & 0.310898 \\
\midrule
\Vertcohirf &   \bfseries{1}   &    \bfseries{1}     &   \bfseries{1}      \\
\bottomrule
\end{tabular}
\end{sc}
\end{small}
\end{center}
\vskip -0.1in
\end{table}

Most collaborative baselines improve performance for one agent but fail to align all agents on a common partition, resulting in asymmetric gains. \DistrKMeans{} reaches a shared consensus, but remains limited by the use of a single $k$-means objective. In contrast, \Vertcohirf{} achieves perfect agreement across agents and fully recovers the global clustering structure, illustrating the benefit of \textit{structural consensus} across heterogeneous modalities. This observation highlights the importance of allowing each agent to select a base clustering method adapted to its own data modality.

\subsection{Influence of the number of agents}
\label{sec:Abline_agent}

We study the impact of the number of collaborative agents on clustering performance in a VFL setting. We consider two variants of \Vertcohirf{}. In \Vertcohirf$\equiv$\Vertcohirf(K-means), all agents use $k$-means as their local base clustering method. In \Vertcohirf(SC-SRGF), all agents rely on the same structure-based local method, SC-SRGF \cite{sc-srgf2020}, which exploits neighborhood and connectivity information rather than Euclidean geometry. We compare against collaborative baselines including \DistrKMeans{}, \Coreset{}, \VPC{}, and \VWay{}.

Experiments are conducted on the {alizadeh-2000-v2} dataset, which contains $n=62$ samples described by $p=2094$ features. This dataset is well suited to vertical feature partitioning, as the feature space can be split across agents while preserving a common sample index.  Features are uniformly distributed among 2 to 6 agents. To reflect realistic partial overlap, each feature is independently shared between any pair of agents with probability $20\%$, with the additional constraint that the overlap between any two agents does not exceed $30\%$ of their feature sets. To account for variability induced by feature partitioning, each experiment is repeated over five independent random splits of the feature space. Reported confidence intervals therefore reflect variability across different feature allocations rather than stochasticity in the clustering algorithms.

\begin{figure}[h]
   \vskip  0.1in
  \begin{center}
    \centerline{\includegraphics[width=\columnwidth]{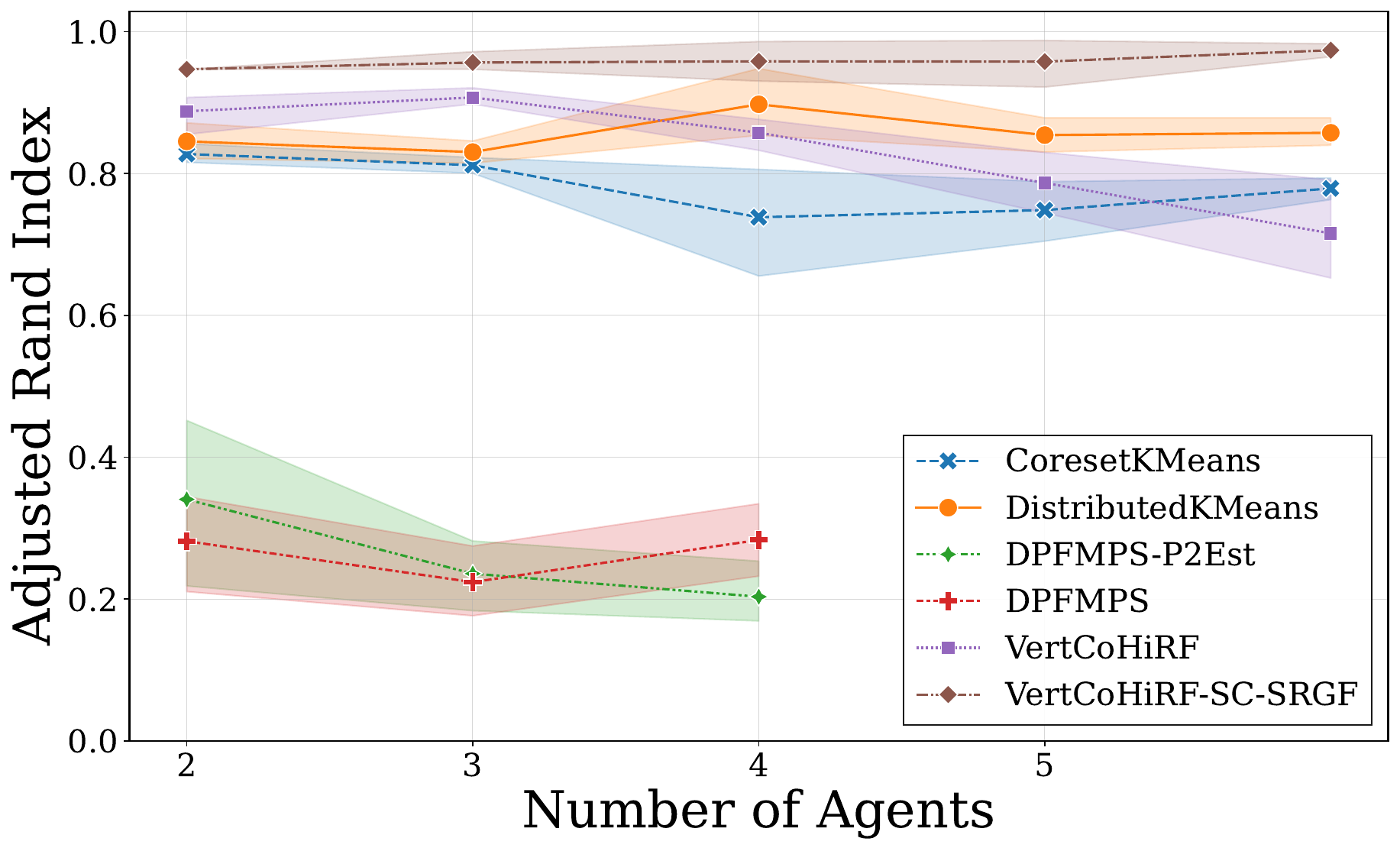}}
    \caption{Clustering performance as a function of the number of collaborative agents, measured by \ARI{}.}
    \label{fig:influence-n-agents}
  \end{center}
\end{figure}

Clustering performance is evaluated using the \ARI{}, which directly measures agreement with the global ground-truth partition and is therefore the most appropriate metric for assessing recovery of the composite clustering structure in this setting. Results are reported in Figure~\ref{fig:influence-n-agents}.

Beyond their inability to scale for more than four agents under our memory constraints, \VPC{} and \VWay{} exhibit substantially lower performance and higher variability as the number of agents increases. By operating on transformed or noisy representations, they struggle to preserve coherent clustering structure under highly heterogeneous views and do not align agents on a common partition in this regime.

\DistrKMeans{} and \Coreset{} enforce a single global $k$-means objective across agents. By imposing a shared Euclidean geometry, these approaches maintain relatively stable performance even as the feature space becomes increasingly fragmented. \Coreset{} nevertheless remains consistently below the best-performing methods due to the information loss induced by summary construction. \DistrKMeans{} performs competitively but is evaluated here in an idealized setting, without privacy constraints and with the true number of clusters provided, and still remains systematically below \Vertcohirf(SC-SRGF). This highlights an intrinsic limitation of enforcing a single global $k$-means objective across heterogeneous feature partitions.


In contrast, \Vertcohirf(K-means) does not impose global geometry, but adopts a different coordination strategy. Each agent independently applies $k$-means on its local features, and coordination is enforced only at the structural level through ordinal consensus. As the number of agents increases, local information becomes more fragmented and the local Euclidean geometry becomes less informative. The consensus mechanism remains effective, but aggregates weaker geometric signals, explaining the gradual decrease in performance as the number of agents grows. 

{This limitation does not apply to \Vertcohirf(SC-SRGF). By relying on structural signals such as neighborhood relations and connectivity, which are less sensitive to feature fragmentation, local structure remains informative even when metric information degrades. As a result, \Vertcohirf(SC-SRGF) maintains high performance and strong stability as the number of agents increases.}


Overall, these results show that in fragmented VFL settings, robustness arises not from enforcing a shared geometric objective or from representation sharing, but from coordinating heterogeneous structures adapted to each local modality.


\subsection{Real world}

For real-world datasets, we summarize clustering performance using \ARI{} boxplots computed over five independent feature partitions. Each boxplot reflects the variability induced by vertical feature splits. The horizontal blue line reports the median \ARI{} obtained by independent local $k$-means without collaboration.

\begin{figure}[h]
\centering

\begin{subfigure}{0.48\linewidth}
\centering
\includegraphics[width=\linewidth]{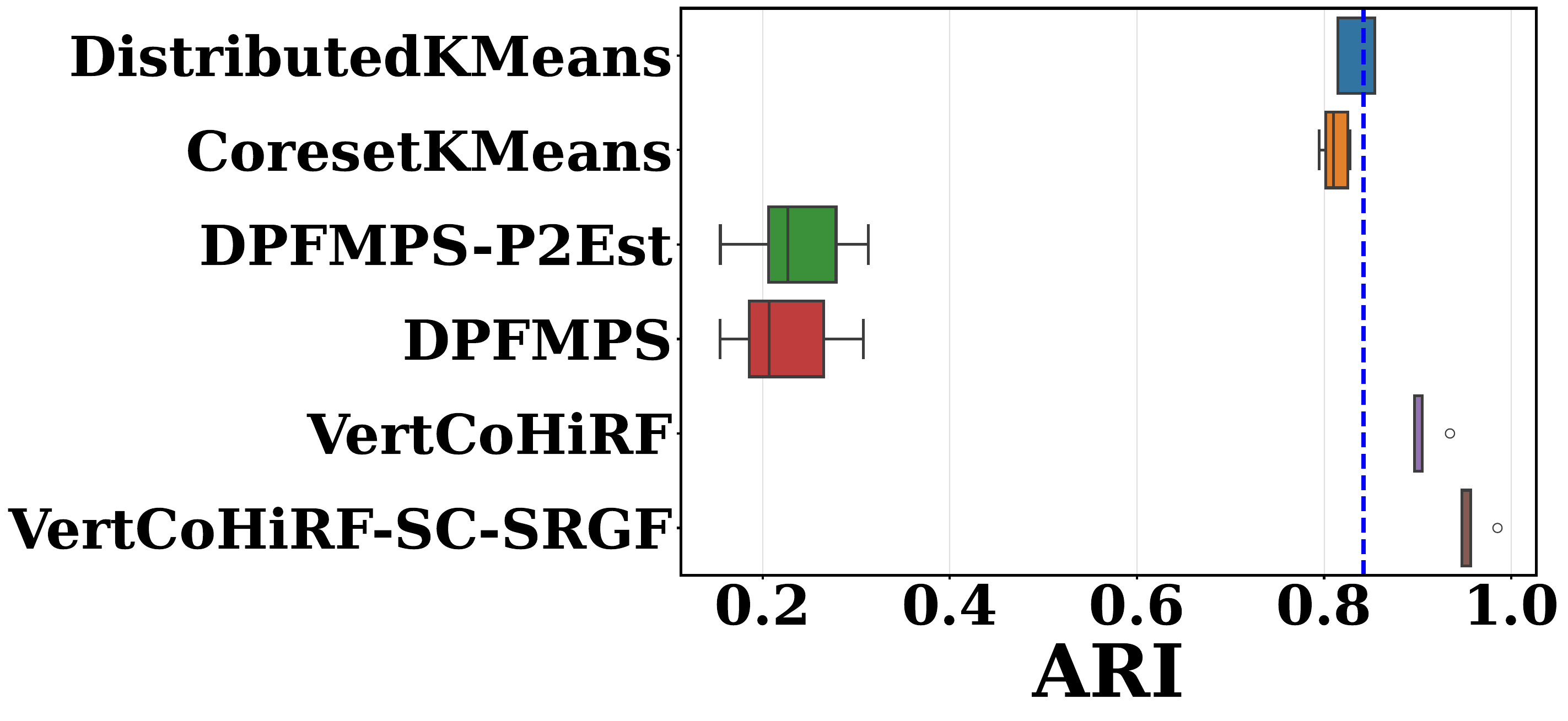}
\caption{alizadeh-2000-v2}
\end{subfigure}
\hfill
\begin{subfigure}{0.48\linewidth}
\centering
\includegraphics[width=\linewidth]{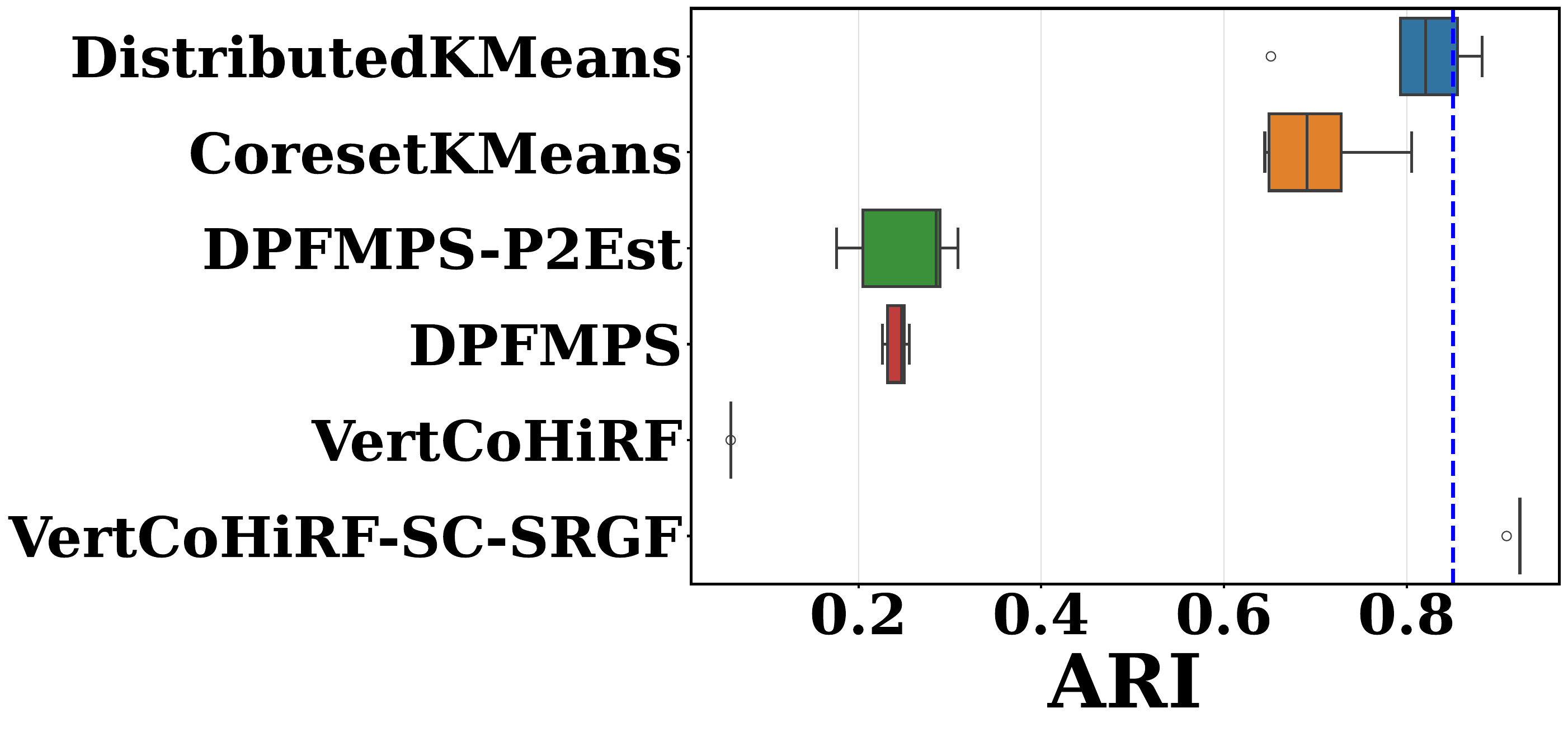}
\caption{chowdary-2006}
\end{subfigure}

\vspace{0.3cm}

\begin{subfigure}{0.48\linewidth}
\centering
\includegraphics[width=\linewidth]{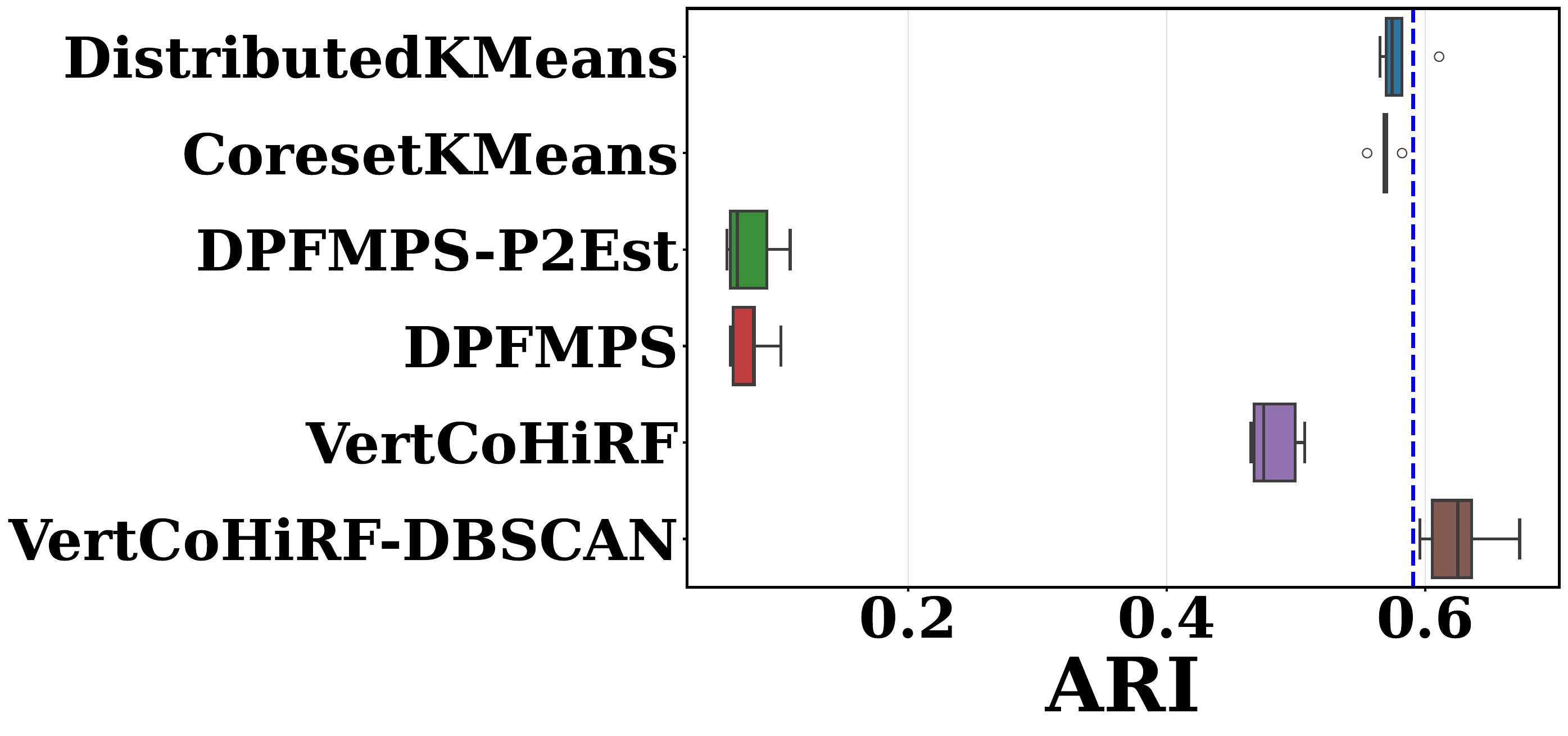}
\caption{coil-20}
\end{subfigure}
\hfill
\begin{subfigure}{0.48\linewidth}
\centering
\includegraphics[width=\linewidth]{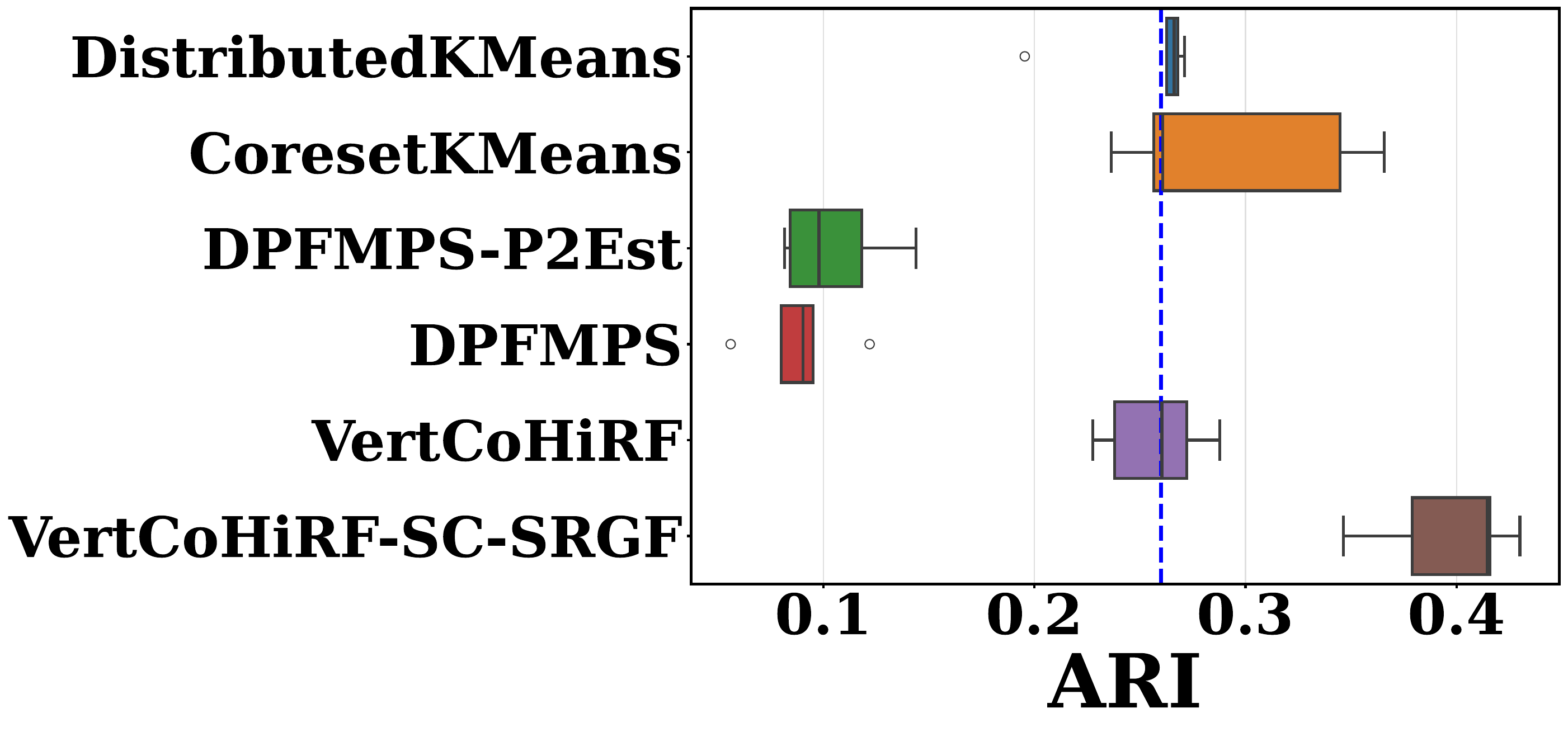}
\caption{garber-2001}
\end{subfigure}

\vspace{0.3cm}

\begin{subfigure}{0.48\linewidth}
\centering
\includegraphics[width=\linewidth]{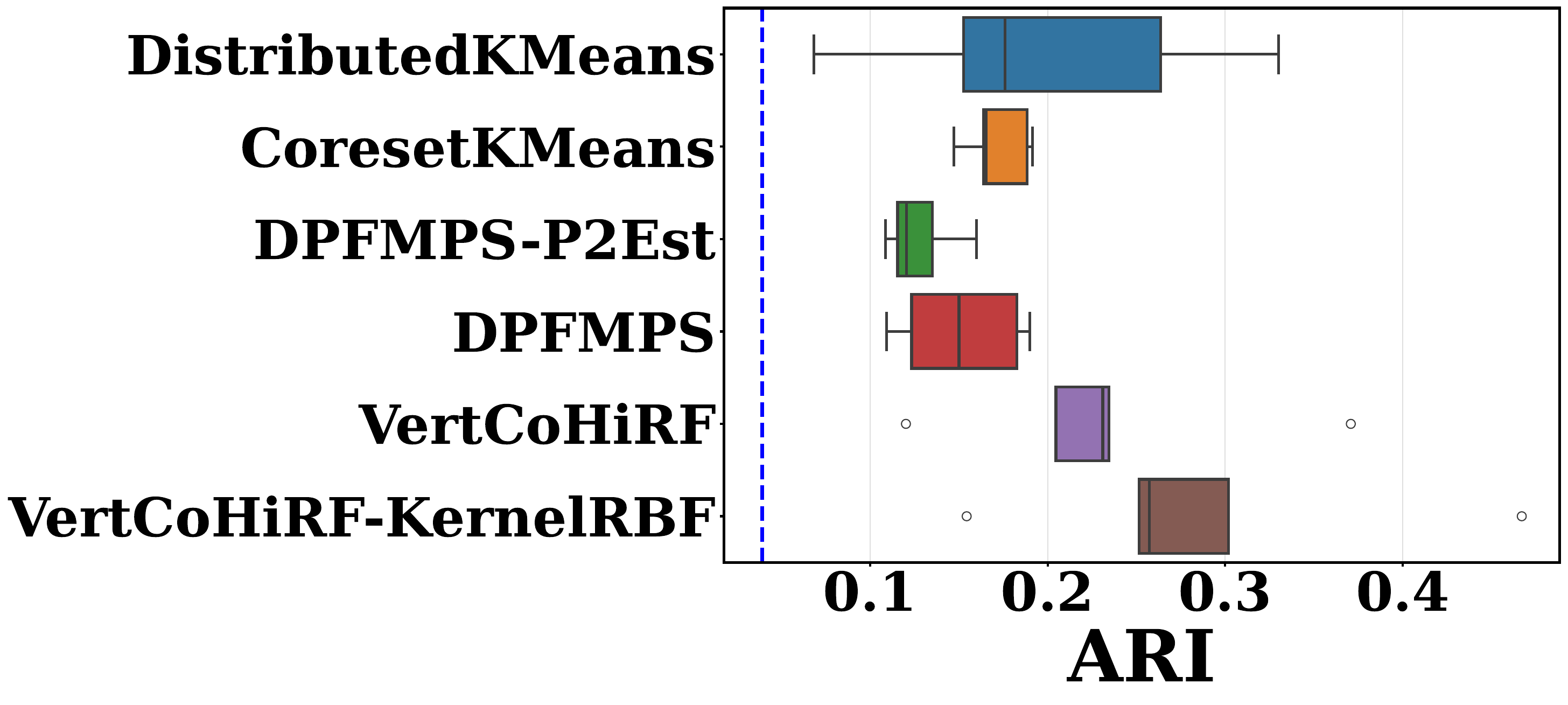}
\caption{nursery}
\end{subfigure}
\hfill
\begin{subfigure}{0.48\linewidth}
\centering
\includegraphics[width=\linewidth]{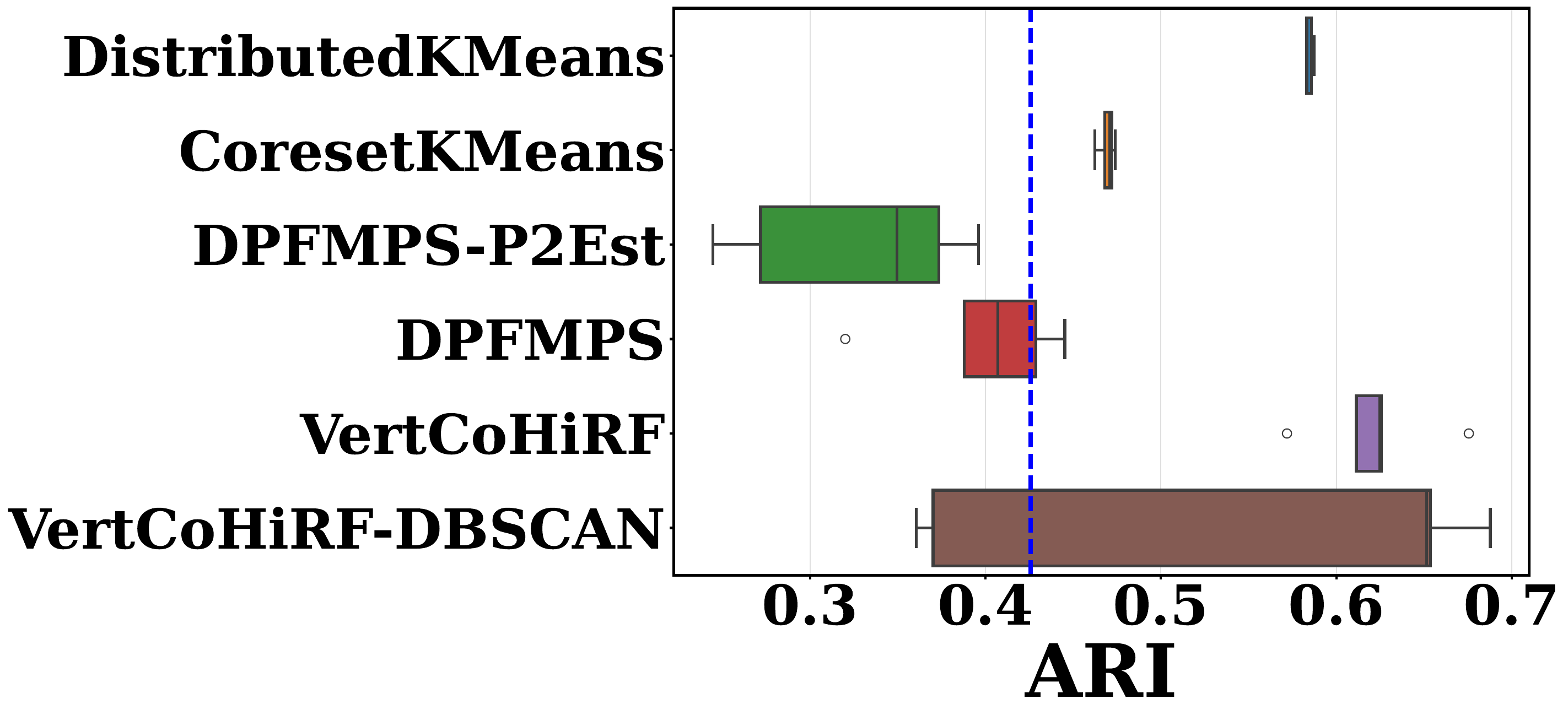}
\caption{shuttle}
\end{subfigure}

\caption{
\ARI{} boxplots across feature partitions. The blue dashed line shows the non-collaborative local $k$-means reference (mean median of agents across five random feature partitions).
}
\label{fig:realdata_ari_boxplots}

\end{figure}

For each dataset, we report only \Vertcohirf{} variants equipped with base clustering methods that are appropriate to the underlying data structure. This choice is fixed \textit{a priori} and applied consistently across all experiments. We additionally include \Vertcohirf(K-means) as a reference, to emphasize that performance gains are not intrinsic to the \Vertcohirf{} framework alone, but depend on the suitability of the chosen local clustering method. Across datasets, we observe that while no single base method is universally optimal, \Vertcohirf{} consistently outperforms collaborative baselines whenever it is paired with a structure-adapted local method. This highlights the importance of \textbf{structural coordination}, rather than enforcing a shared objective across agents, in VFL settings.

Although \VPC{} and \VWay{} consistently underperform in terms of \ARI{}, this behavior is primarily due to the noise injection and representation sharing mechanisms required to ensure DP, rather than to an intrinsic limitation of the methods themselves. They therefore serve as informative differentially private baselines, highlighting the accuracy loss induced by privacy-preserving noise. Full numerical results and additional metrics are reported in Appendix~\ref{app=experiment}.

\section{Discussion and limitations}


{\Vertcohirf{} is designed to extract shared structural information across heterogeneous feature views when such structure exists. Accordingly, its benefits are most pronounced when multiple agents provide complementary and informative perspectives on the same samples. When only a single view is informative, or when the structures induced by different views are largely incompatible, the consensus mechanism may yield limited additional gains for the strongest individual agent. In these regimes, the strongest agent may still guide or stabilize weaker agents, while its own structure can be conservatively constrained by cross-view disagreement. Crucially, the veto-based consensus mechanism prevents the introduction of spurious groupings by allowing agreement only when a shared structure is supported across views. In addition, \Vertcohirf{} remains a safe and interpretable aggregation mechanism (CFH), preserving structural validity while enabling meaningful collaboration whenever complementary information is present. Future work will investigate the adaptation of our approach to hybrid partitioning regimes.
}

\newpage
\section*{Impact Statement}

This paper presents work whose goal is to advance the field of 
Machine Learning. There are many potential societal consequences 
of our work, none which we feel must be specifically highlighted here.


\bibliography{ref}
\bibliographystyle{icml2026}

\newpage
\appendix
\onecolumn

\section{Appendix: Detailed theoretical analysis}
\label{app:theory}

\subsection{Communication complexity}
\label{app:comm}

We derive the communication complexity of \Vertcohirf{} by analyzing separately
the two consensus phases executed at each iteration.

\subsubsection{Cluster consensus phase}

At iteration $e$, each agent $a \in [A]$ computes a local cluster assignment vector
$$
\L_a^{(e)} \in \{1,\ldots,C_a\}^{n^{(e-1)}},
$$
defined over the current set of $n^{(e-1)}$ active medoids. This vector is broadcast to the remaining $A-1$ agents.

Each cluster label requires $\lceil \log_2(C_a) \rceil$ bits for binary encoding.
Hence, the number of bits transmitted by agent $a$ during this phase is
\begin{equation}
    (A-1)\, n^{(e-1)} \, \lceil \log_2(C_a) \rceil.
\end{equation}

Summing over all agents yields
\begin{equation}
b^{(e)}_{\mathrm{cluster}}=
\sum_{a=1}^A (A-1)\, n^{(e-1)} \, \lceil \log_2(C_a) \rceil
\leq
A(A-1)\, n^{(e-1)} \, \lceil \log_2(\max_a C_a) \rceil.
\end{equation}

\subsubsection{Medoid consensus phase}

Let $n^{(e)}$ denote the number of consensual clusters formed at iteration $e$.
For each such cluster, every agent transmits a ranked list of at most $N_s$
candidate medoids, expressed as global sample identifiers.

Each identifier requires $\lceil \log_2(n) \rceil$ bits for binary encoding,
where $n$ denotes the total number of samples. The number of bits transmitted by
one agent during this phase is therefore at most
\begin{equation}
    (A-1)\, n^{(e)}\, N_s \, \lceil \log_2(n) \rceil.
\end{equation}

Aggregating over all agents yields
\begin{equation}
    b^{(e)}_{\mathrm{medoid}}
    \leq
    A(A-1)\, n^{(e)}\, N_s \, \lceil \log_2(n) \rceil.
\end{equation}

\subsubsection{Total communication}

Combining the two phases yields the stated bound:
\begin{equation}
b^{(e)}
\leq 
A(A-1)\left[
n^{(e-1)} \lceil \log_2(\max_a C_a) \rceil
+
n^{(e)} S \lceil \log_2(n) \rceil
\right].
\end{equation}

Since $n^{(e)}$ decreases monotonically across iterations, the overall
communication cost is dominated by the first iterations and scales linearly
with the number of samples, up to logarithmic factors.

\subsection{Proof of Identifier-Level Structural Privacy (ILSP)}
\label{app:privacy}

At each iteration of \Vertcohirf{}, agents exchange only: (i) cluster assignment vectors $\L_a^{(e)}$, (ii) ranked lists of sample identifiers during the medoid consensus phase.

By construction, these messages depend on an agent's feature matrix only through the induced cluster assignments and ordinal rankings. No feature-dependent numerical values are transmitted. Therefore, for any agent $b \neq a$, and for any two feature matrices
$\bbX_b$ and $\bbX_b'$ that induce identical cluster assignments and identical rankings at all iterations, the complete message history received by agent $a$ is identical under both configurations. This implies that
$$
\mathbb{P}\left(M_a^{(1:E)} \mid \bbX_b\right)
=
\mathbb{P}\left(M_a^{(1:E)} \mid \bbX_b'\right),
$$
which establishes identifier-level structural privacy.

\subsection{Structural guarantees under strict consensus}
\label{app:theory_struct}


Consider two distinct points $m_1$ and $m_2$. By definition of strict consensus, $m_1$ and $m_2$ belong to the same consensus cluster if and only if, for every agent $a$, the labels assigned to $m_1$ and $m_2$ by $a$ are identical. Assume that there exists at least one honest agent $h$ such that $h$ assigns different labels to $m_1$ and $m_2$. Then the strict consensus condition is not satisfied, and $m_1$ and $m_2$ cannot belong to the same consensus cluster, regardless of the behavior of the other agents. It follows that no fusion between two points distinguished by at least one honest agent can be imposed by Byzantine agents. Consequently, any neighborhood relation present in a consensus cluster produced by Phase~1 is necessarily compatible with the partitions produced by all honest agents. Byzantine agents can only block certain merges by introducing additional disagreements, which may increase fragmentation or slow down convergence, but cannot create structurally incorrect groupings.

\subsection{Robustness to Byzantine ranking attacks}
\label{app:attack}

We design a controlled synthetic experiment to empirically study the robustness of the medoid selection phase of \Vertcohirf{} under a single Byzantine attacker. The goal is to validate the theoretical analysis of Phase~2 by isolating the effect of adversarial ordinal manipulation as a function of the signal-to-noise ratio, the results are presented on \cref{fig:attack}.


\subsubsection{Protocole}

We generate a dataset with $n=1000$ samples drawn from $C=4$ clusters in a low-dimensional continuous space. Each cluster contains $n/C$ points and is generated from an isotropic Gaussian distribution with fixed centers. The standard deviation $\sigma$ of the Gaussians is varied in $(0,1]$. Increasing $\sigma$  progressively degrades cluster separability, eventually making the latent structure indistinguishable from noise.

The feature space is vertically partitioned across $A=3$ agents. Informative features are evenly split among agents, and additional noise features are injected so that each agent observes only a partial and noisy view of the data. There is no feature overlap between agents. All agents share the same sample index set.

The experiment \textbf{focuses exclusively on Phase~2 of \Vertcohirf{}. No attack is performed during Phase~1}: all agents, including the Byzantine one, operate on the same candidate set, which is assumed to have been validated by unanimous consensus. This isolates the robustness of the medoid selection mechanism. Two agents behave honestly. A single agent acts as a Byzantine adversary. The adversary do not observe the behavior of the honest agents. Its attack is purely ordinal: the adversarial agent computes a local ranking based on its own feature view and submits \textbf{an adversarial permutation} of this ranking during the medoid selection step, with the goal of promoting a less representative candidate.

For each value of $\sigma$ , the experiment is repeated over multiple random trials. Performance is measured using the \ARI{}, averaged over runs, with confidence intervals reflecting variability across trials.

\subsubsection{Results and discussion}


The Figure~\ref{fig:attack} reports the \ARI{} obtained by \Vertcohirf{} with and without an attacker as a function of the cluster standard deviation $\sigma$ . For small values of $\sigma$ , clusters are well separated and honest agents exhibit strong ordinal agreement. In this regime, the performance gap between the attacked and non-attacked settings remains limited. This empirically confirms that when the aggregated honest signal dominates the adversarial capacity, ordinal manipulation by a single Byzantine agent cannot significantly alter the medoid selection outcome.

As $\sigma$  increases, cluster separability deteriorates and the consensus quality decreases. Consequently, the \ARI{} smoothly degrades in both settings. The attacked curve consistently lies below the non-attacked one, but the gap remains controlled and does not lead to catastrophic failure. This behavior is consistent with the theoretical prediction that attacks become effective only when the honest signal-to-noise ratio collapses.

In the extreme regime $\sigma \approx 1$, the signal is almost entirely drowned in noise. Honest rankings become nearly random. In this limit, the \ARI{} approaches zero regardless of the presence of an attacker, reflecting a fundamental information-theoretic limit rather than a failure of the protocol.


This experiment provides a concrete illustration of the robustness guarantees established in the theoretical analysis. It shows that a Byzantine agent restricted to ordinal manipulation and operating without coordination cannot dominate medoid selection as long as a minimal ordinal consensus exists among honest agents. Performance degradation occurs smoothly as the signal weakens and does not result in abrupt failure, confirming both the central role of the first iteration and the progressive strengthening of robustness as the algorithm advances.

\section{Additional experimental results}
\label{app=experiment}

We examine the impact of collaboration on five real-world datasets, which are characterized in \Cref{tab:real_datasets_characteristics} by the number of samples $n$, the number of features $p$, the number of clusters $C$, and the number of categorical features $p_{cat}$. All datasets are accessible via the OpenML platform \cite{openml2020}.

The selected datasets encompass diverse configurations in terms of sample size, dimensionality, feature types, and cluster geometry. In particular, the alizadeh-2000-v2 and garber-2001 datasets correspond to high-dimensional scenarios in which the number of features substantially exceeds the number of samples ($p \gg n$). Such settings are characteristic of bioinformatics applications and are especially appropriate for evaluating collaborative methods when the feature space is distributed across multiple agents. The chowdary-2006 dataset represents an intermediate regime where $p \approx n$. The coil-20 dataset comprises image data, exhibiting a high-dimensional feature space derived directly from pixel intensities. The nursery dataset is originally categorical and has been transformed via one-hot encoding, yielding a low intrinsic (pre-encoding) dimensionality but a purely categorical feature space with a total of 19 features after encoding. Finally, the shuttle dataset exemplifies a large-scale setting with $n \gg p$.

For all real-world datasets, experiments are repeated over five independent random partitions of the feature space. Confidence intervals therefore capture variability induced by feature allocation across agents, rather than stochastic effects of the clustering algorithms.


\begin{table}[h]
\caption{Characteristics of real-world datasets used in our experiments.}
\label{tab:real_datasets_characteristics}
\begin{center}
\begin{small}
\begin{sc}
\begin{tabular}{lrrrrr}
\toprule
Dataset & OpenML ID & $n$ & $p$ & $C$ & $p_{cat}$ \\
\midrule
alizadeh-2000-v2 & 46773 & 62 & 2094 & 3 & 0 \\
chowdary-2006 & 46778 & 104 & 183 & 2 & 0 \\
coil-20 & 46783 & 1440 & 1025 & 20 & 0 \\
garber-2001 & 46779 & 66 & 4554 & 4 & 0 \\
nursery & 1568 & 12958 & 9 & 4 & 9 \\
shuttle & 40685 & 58000 & 10 & 7 & 1 \\
\bottomrule
\end{tabular}
\end{sc}
\end{small}
\end{center}
\vskip -0.1in
\end{table}

\subsection{Local relax consensus ($^*$ variants)}


Variants marked with a superscript $^*$ correspond to a relaxed local consensus
strategy. {See \cite{Cohirf} for details on these variants}. In these variants, each agent performs an additional local filtering step before consensus, discarding projections or views that are inconsistent with the majority of its local representations. This mechanism improves robustness when local projections are noisy or weakly informative, and is applied prior to the global inter-agent consensus.

For SC-SRGF, each agent performs a single local projection ($R_a = 1$) using
all available features ($q_a = p_a$). As this method directly operates in the
original feature space and exploits neighborhood and connectivity information,
dimensionality reduction and relax consensus are not required. Consequently,
no $^*$ variant is considered for SC-SRGF.

\subsection{Hyperparameter optimization}

As detailed in \cref{sec:Experiments}, we perform hyperparameter optimization to ensure that each model is assessed under its optimal configuration. The corresponding search space for each algorithm and for the Base Clustering Methods used with \Vertcohirf{} are summarized in \cref{tab:hpo-search-space,tab:bcm-hpo-search-space}. For each experiment, we optimize the associated evaluation metric over 50 trials using Optuna \cite{akibaOptunaNextgenerationHyperparameter2019} in combination with the Tree-structured Parzen Estimator (TPE) sampler.

\begin{table}[h]
\centering
\caption{Hyperparameter search spaces for every model used in our experiments.}
\label{tab:hpo-search-space}
\begin{tabular}{@{}lll@{}}
\toprule
Model                                   & Parameter                          & Values          \\ \midrule
\multirow{3}{*}{VertCoHiRF}             & Percentage of features ($q_a{\%}$) & Float[0.1, 1.0] \\
                                        & Repetitions ($R_a$)                & Int[2, 10]      \\
                                        & Relaxed Threshold (h)              & 0.8             \\ \midrule
\DistrKMeans / \Coreset / \VPC / \VWay & Number of clusters                 & Int[2, 30]      \\ \midrule
\multicolumn{3}{l}{\parbox{12cm}{\footnotesize $^a$ All VertCoHiRF variants use the same hyperparameter search space as their base clustering methods, plus the listed VertCoHiRF-specific hyperparameters, except CoHiRF-SC-SRGF, where the number of repetitions and the percentage of features are implicitly determined by the number of affinity matrices and sampling ratios and are therefore fixed to 1 and 1.0, respectively.}} \\ \bottomrule
\end{tabular}
\end{table}

\begin{table}[]
\centering
\caption{Hyperparameter search spaces for every Base Clustering Method (BCM) used in our experiments.}
\label{tab:bcm-hpo-search-space}
\begin{tabular}{@{}lll@{}}
\toprule
Model                            & Parameter                   & Values           \\ \midrule
\multirow{2}{*}{DBSCAN}          & $\epsilon$                  & Float[0.1, 10.0] \\
                                 & Minimum number of samples   & Int[2, 50]       \\ \midrule
KMeans                           & Number of clusters          & Int[2, 30]       \\ \midrule
\multirow{2}{*}{KernelRBFKMeans} & Number of clusters          & Int[2, 30]       \\
                                 & $\gamma$                    & Float[0.1, 30.0] \\ \midrule
\multirow{3}{*}{SC-SRGF}         & Number of affinity matrices & Int[10, 30]      \\
                                 & Sampling ratio              & Float[0.2, 0.8]  \\
                                 & Number of clusters          & Int[2, 30]       \\ \bottomrule
\end{tabular}
\end{table}




\subsection{Real World datasets silhouette}


In addition to ARI, we report Silhouette scores as an internal validation metric. These results are provided for completeness, but should be interpreted with caution in vertically federated settings.

\begin{figure*}[h]
\vskip 0.1in
\centering
\begin{subfigure}[b]{0.32\textwidth}
\centering
\includegraphics[width=\linewidth]{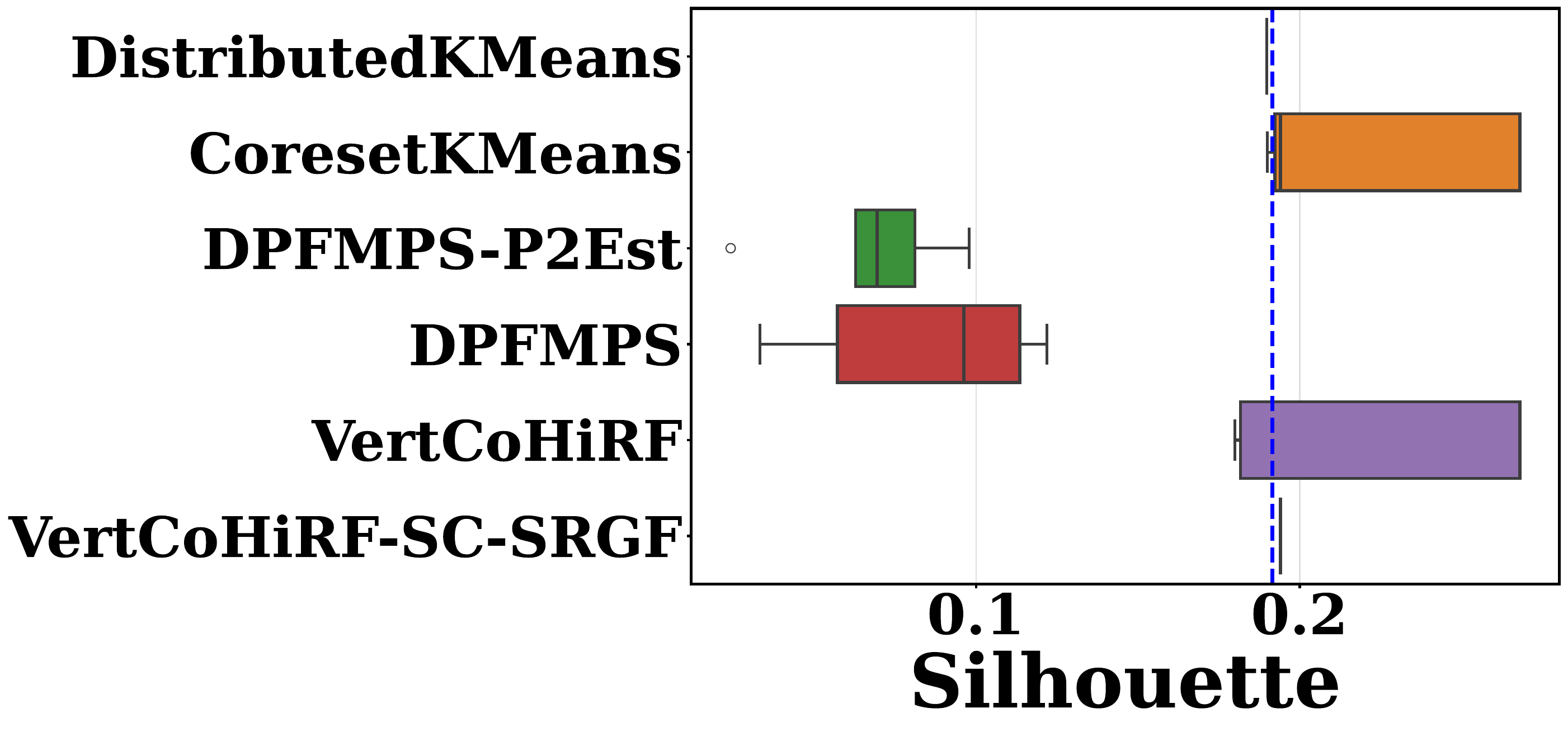}
\caption{{alizadeh-2000-v2}}
\end{subfigure}
\hfill
\begin{subfigure}[b]{0.32\textwidth}
\centering
\includegraphics[width=\linewidth]{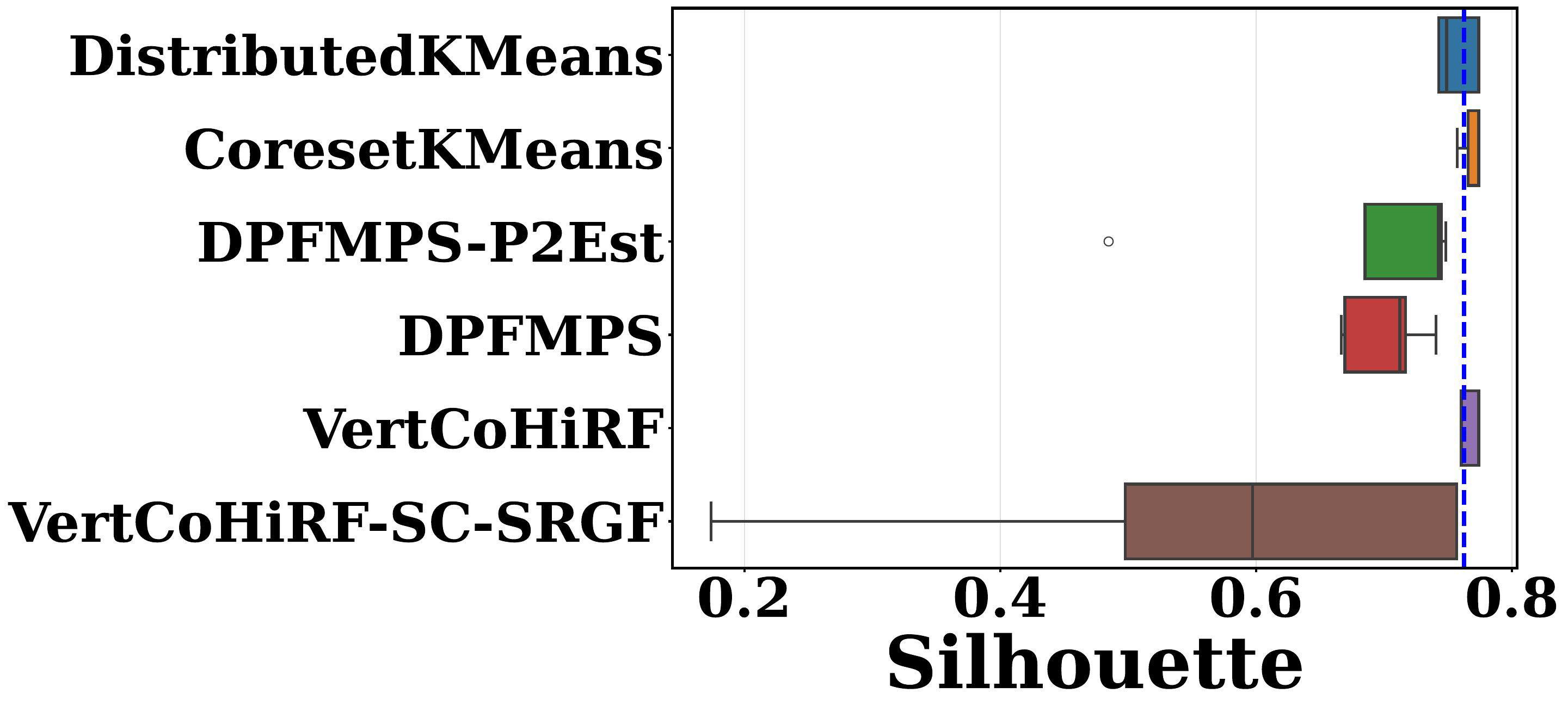}
\caption{{chowdary-2006}}
\end{subfigure}
\hfill
\begin{subfigure}[b]{0.32\textwidth}
\centering
\includegraphics[width=\linewidth]{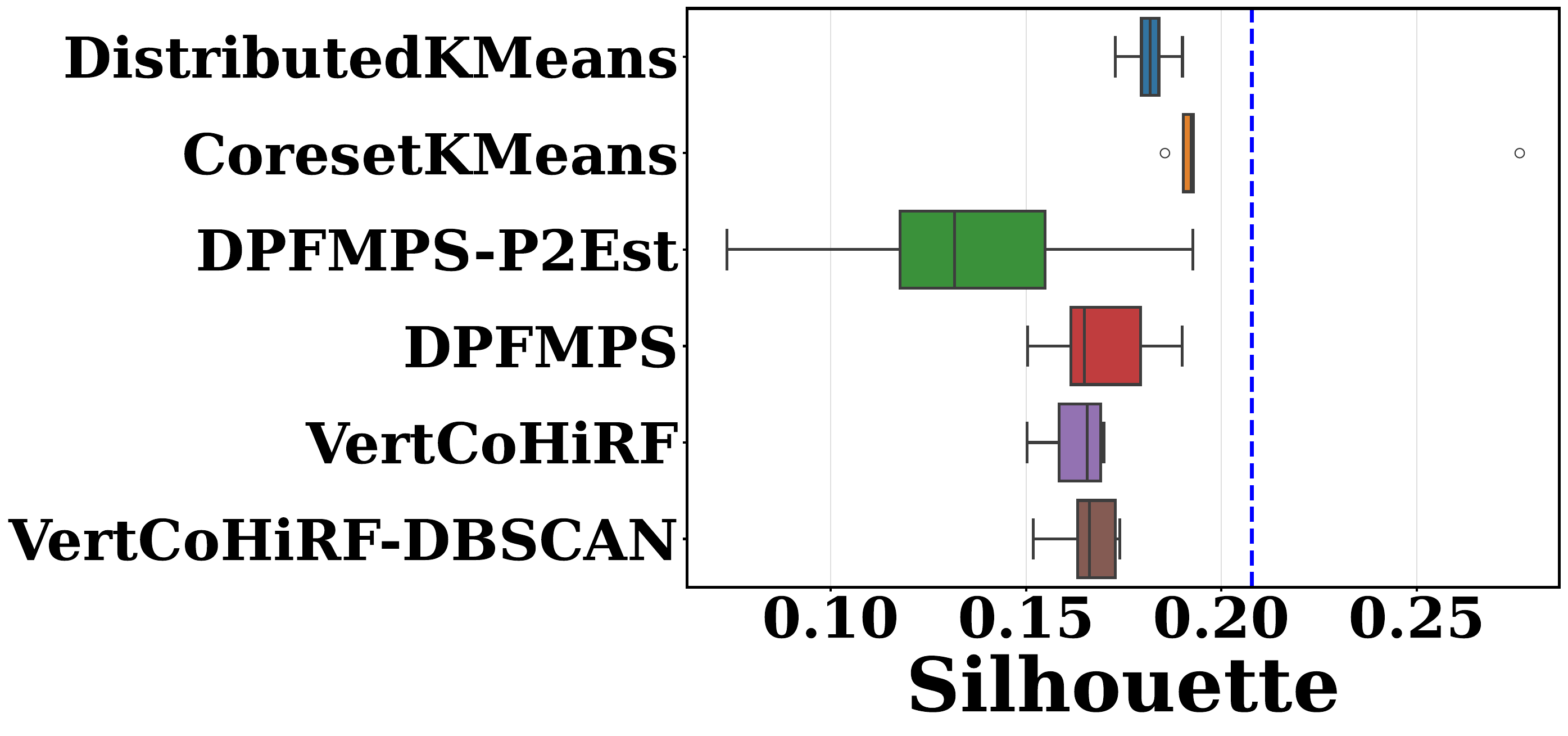}
\caption{{coil-20}}
\end{subfigure}

\vskip 0.15in

\begin{subfigure}[b]{0.32\textwidth}
\centering
\includegraphics[width=\linewidth]{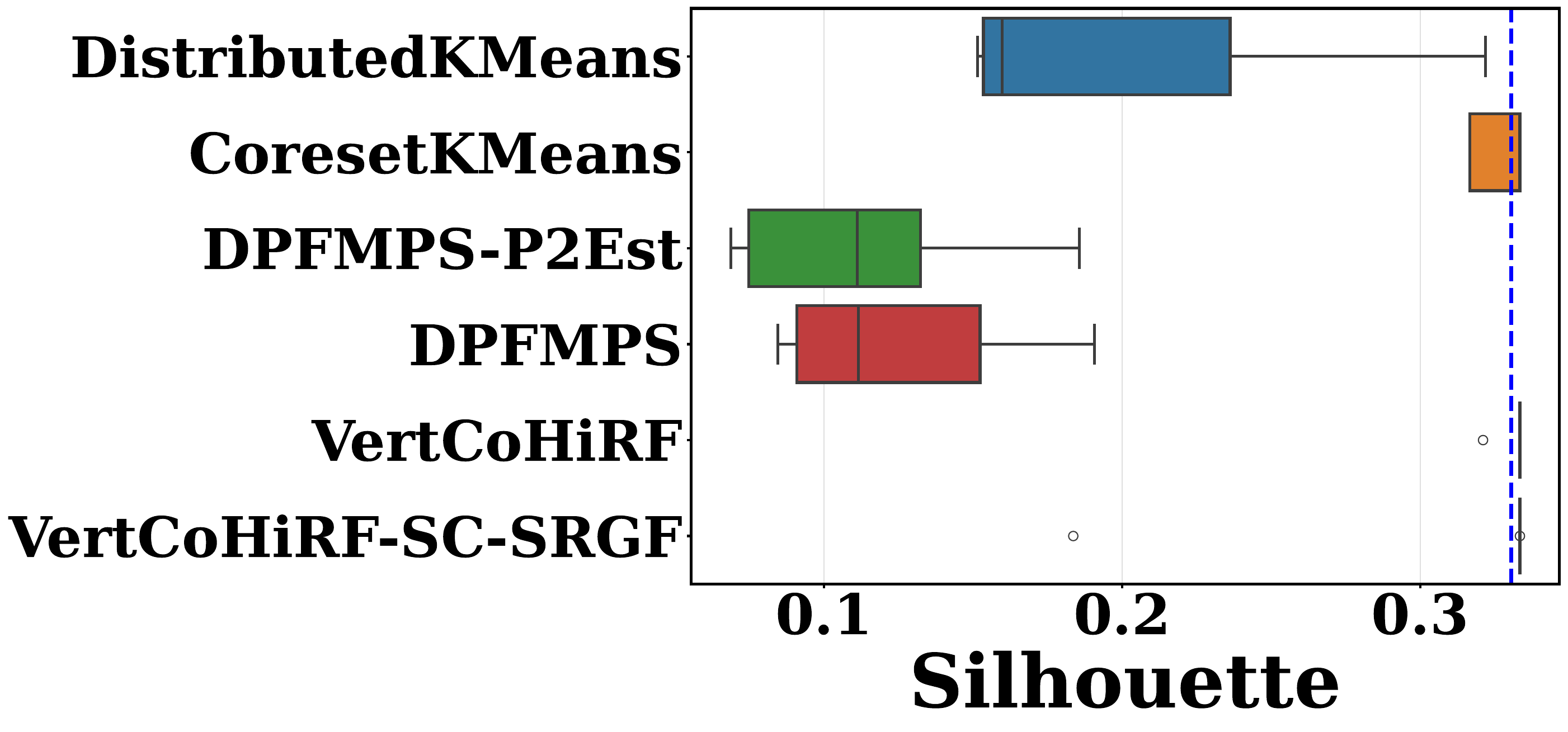}
\caption{{garber-2001}}
\end{subfigure}
\hfill
\begin{subfigure}[b]{0.32\textwidth}
\centering
\includegraphics[width=\linewidth]{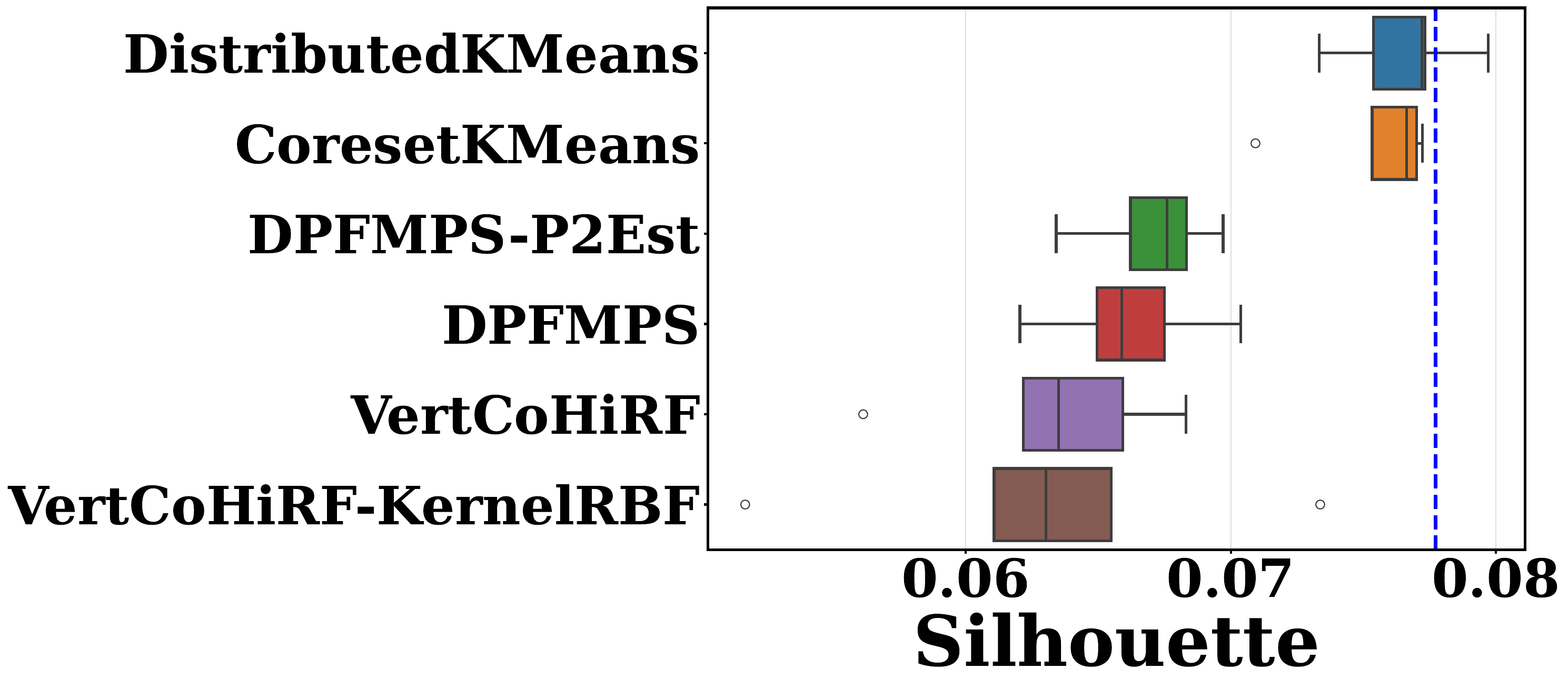}
\caption{{nursery}}
\end{subfigure}
\hfill
\begin{subfigure}[b]{0.32\textwidth}
\centering
\includegraphics[width=\linewidth]{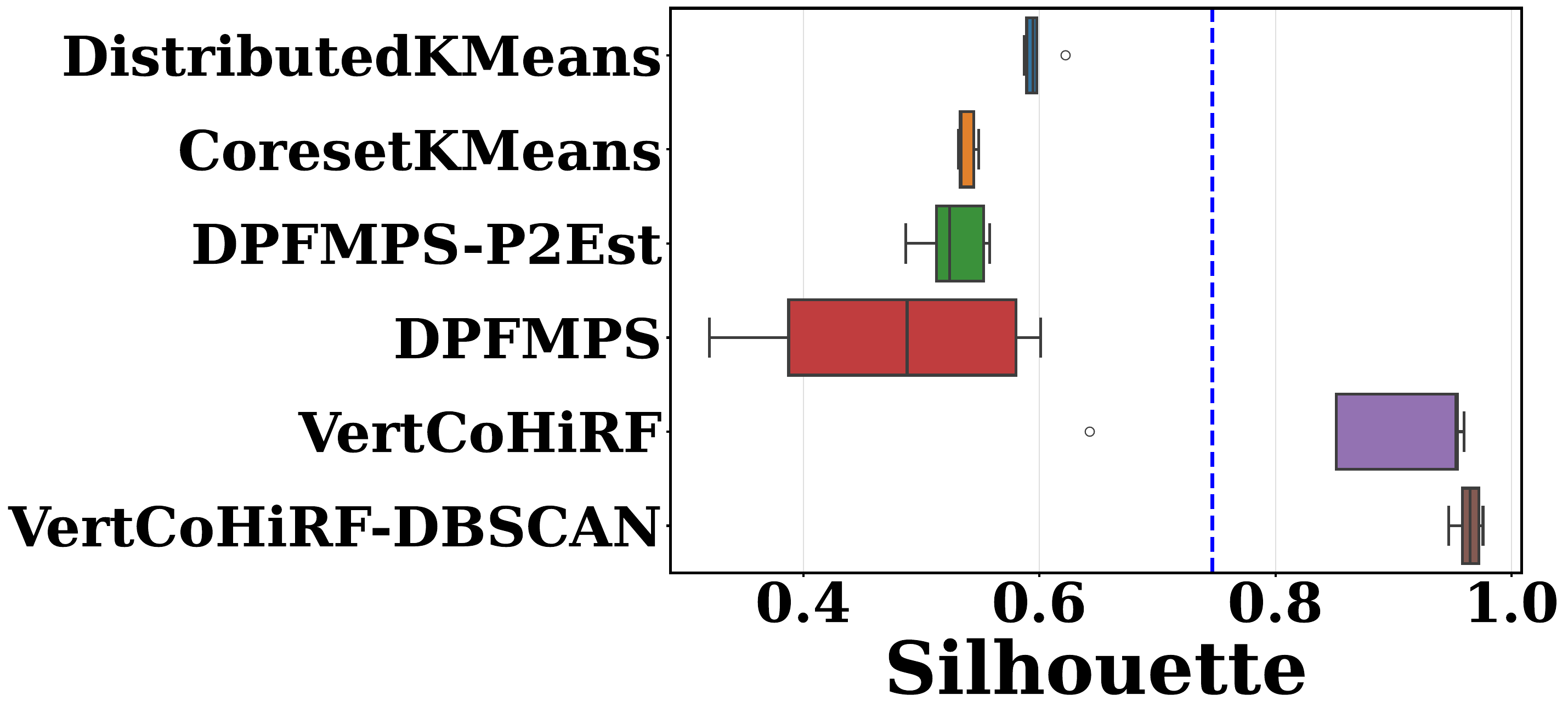}
\caption{{shuttle}}
\end{subfigure}
\caption{
Silhouette boxplots across feature partitions. The blue dashed line  shows the non-collaborative local $k$-means reference (mean median of agents across five random feature partitions ).
}
\label{fig:realdata_sil_boxplots}
\end{figure*}

The Silhouette score is an internal validation metric based on within-cluster compactness and between-cluster separation under a single Euclidean metric. In vertically federated learning, each agent observes only a partial and heterogeneous subset of features, and no agent has access to the full geometry of the data. As a consequence, Silhouette scores computed after collaboration may not reflect the quality of the recovered global clustering structure, even when external agreement with ground truth is high.

This effect is particularly pronounced in multimodal or heterogeneous datasets, where different feature subsets encode complementary but non-aligned structures. In such cases, external metrics such as ARI provide a more faithful evaluation of whether the composite clustering structure has been correctly recovered.

\subsection{Complete results tables}


These tables report the complete quantitative results for all datasets, including both external (ARI) and internal (Silhouette) validation metrics. Results are reported as mean $\pm$ standard deviation over five independent random feature partitions. {We denote by KK , the kernel K-Means  where an RBF kernel feature map is approximated using 500 random Fourier features.}

\begin{table}[h]
\centering
\begin{minipage}{0.48\linewidth}
\centering
\caption{Clustering results on {alizadeh-2000-v2} (ARI)}
\label{tab:Alizadeh_ARI}
\begin{footnotesize}
\begin{sc}
\begin{tabular}{lll}
\toprule
Model & Collab. ARI & Diff. ARI \\
\textcolor{blue}{Ref: K-Means}  & \textcolor{blue}{0.834 $\pm$ 0.018} & \textcolor{blue}{0.000 $\pm$ 0.000} \\
\midrule
\DistrKMeans  & 0.830 $\pm$ 0.021 & -0.003 $\pm$ 0.033 \\
\Coreset      & 0.812 $\pm$ 0.014 & -0.022 $\pm$ 0.022 \\
\VPC          & 0.224 $\pm$ 0.062 & -0.610 $\pm$ 0.073 \\
\VWay         & 0.236 $\pm$ 0.062 & -0.598 $\pm$ 0.066 \\
\midrule
\Vertcohirf$^*$(Kmeans) & \underline{0.907 $\pm$ 0.016} & \underline{0.074 $\pm$ 0.023} \\
\Vertcohirf(SC-SRGF) & \bfseries 0.957 $\pm$ 0.016 & \bfseries 0.123 $\pm$ 0.031 \\
\bottomrule
\end{tabular}
\end{sc}
\end{footnotesize}
\end{minipage}
\hfill
\begin{minipage}{0.48\linewidth}
\centering
\caption{Clustering results on {alizadeh-2000-v2} (Silhouette)}
\label{tab:Alizadeh_Sil}
\begin{footnotesize}
\begin{sc}
\begin{tabular}{|lll}
\toprule
Model & Collab. Sil & Diff. Sil \\
\textcolor{blue}{Ref: K-Means} & \textcolor{blue}{0.190 $\pm$ 0.000} & \textcolor{blue}{0.000 $\pm$ 0.000} \\
\midrule
\DistrKMeans & 0.190 $\pm$ 0.000 & 0.000 $\pm$ 0.000 \\
\Coreset & \bfseries 0.222 $\pm$ 0.042 & \bfseries 0.033 $\pm$ 0.042 \\
\VPC & 0.084 $\pm$ 0.038 & -0.105 $\pm$ 0.038 \\
\VWay & 0.067 $\pm$ 0.027 & -0.123 $\pm$ 0.027 \\
\midrule
\Vertcohirf$^*$(Kmeans) & \underline{0.216 $\pm$ 0.048} & \underline{0.026 $\pm$ 0.048} \\
\Vertcohirf(SC-SRGF) & 0.194 $\pm$ 0.000 & 0.004 $\pm$ 0.000 \\
\bottomrule
\end{tabular}
\end{sc}
\end{footnotesize}
\end{minipage}
\end{table}


\begin{table}[h]
\centering
\begin{minipage}{0.48\linewidth}
\centering
\caption{Clustering results on dataset {chowdary-2006}}
\label{tab:Chowdary_ARI}
\begin{footnotesize}
\begin{sc}
\begin{tabular}{lll}
\toprule
Model & Collab. ARI & Diff. ARI \\
\textcolor{blue}{Ref: K-Means} & \textcolor{blue}{0.841 $\pm$ 0.020}&  \, \textcolor{blue}{0.000 $\pm$ 0.000} \\
\midrule
\DistrKMeans  & \underline{0.801 $\pm$ 0.090} & \underline{-0.041 $\pm$ 0.087} \\
\Coreset  & 0.704 $\pm$ 0.066 & -0.138 $\pm$ 0.084 \\
\VPC  & 0.242 $\pm$ 0.012 & -0.599 $\pm$ 0.024 \\
\VWay  & 0.253 $\pm$ 0.059 & -0.588 $\pm$ 0.049 \\
\midrule
\Vertcohirf$^*$(Kmeans)  & 0.060 $\pm$ 0.000 & -0.781 $\pm$ 0.020 \\
\Vertcohirf(SC-SRGF)  & \bfseries 0.921 $\pm$ 0.007 &\,\bfseries  0.080 $\pm$ 0.023 \\
\bottomrule
\end{tabular}
\end{sc}
\end{footnotesize}
\end{minipage}
\hfill
\begin{minipage}{0.48\linewidth}
\centering
\caption{Clustering results on dataset {chowdary-2006}}
\label{tab:Chowdary_Sil}
\begin{footnotesize}
\begin{sc}
\begin{tabular}{|lll}
\toprule
Model &  Collab. Sil& Diff. Sil\\
\textcolor{blue}{Ref: K-Means} & \textcolor{blue}{0.767 $\pm$ 0.009}&  \, \textcolor{blue}{0.000 $\pm$ 0.000} \\
\midrule
\DistrKMeans  & 0.756 $\pm$ 0.016 & -0.011 $\pm$ 0.013 \\
\Coreset  & \bfseries 0.769 $\pm$ 0.007 &\,0.002 $\pm$ 0.009 \\
\VPC  & 0.701 $\pm$ 0.032 & -0.066 $\pm$ 0.040 \\
\VWay  & 0.681 $\pm$ 0.113 & -0.086 $\pm$ 0.116 \\
\midrule
\Vertcohirf$^*$(Kmeans)  & \bfseries 0.769 $\pm$ 0.007 &\,\bfseries 0.001 $\pm$ 0.011 \\
\Vertcohirf(SC-SRGF)  & 0.556 $\pm$ 0.241 & -0.211 $\pm$ 0.243 \\
\bottomrule
\end{tabular}
\end{sc}
\end{footnotesize}
\end{minipage}
\end{table}



\begin{table}[h]
\centering
\begin{minipage}{0.48\linewidth}
\centering

\caption{Clustering results on dataset coil-20}
\label{tab:Coil_ARI}
\begin{footnotesize}
\begin{sc}
\begin{tabular}{lll}
\toprule
Model & Collab. ARI & Diff. ARI \\
\textcolor{blue}{Ref: K-Means} & \textcolor{blue}{0.588 $\pm$ 0.014} &  \, \textcolor{blue}{0.000 $\pm$ 0.000} \\
\midrule
\DistrKMeans  & 0.580 $\pm$ 0.018 & -0.008 $\pm$ 0.010 \\
\Coreset  & 0.569 $\pm$ 0.010 & -0.019 $\pm$ 0.023 \\
\VPC  & 0.078 $\pm$ 0.016 & -0.510 $\pm$ 0.014 \\
\VWay  & 0.078 $\pm$ 0.021 & -0.510 $\pm$ 0.021 \\
\midrule
\Vertcohirf(Kmeans)  & 0.483 $\pm$ 0.019 & -0.105 $\pm$ 0.025 \\
\Vertcohirf$^*$(DBSCAN)  &\bfseries  0.627 $\pm$ 0.030 &\,\bfseries 0.039 $\pm$ 0.041 \\
\bottomrule
\end{tabular}
\end{sc}
\end{footnotesize}

\end{minipage}
\hfill
\begin{minipage}{0.48\linewidth}
\centering

\caption{Clustering results on dataset coil-20}
\label{tab:Coil_Sil}
\begin{footnotesize}
\begin{sc}
\begin{tabular}{|lll}
\toprule
Model &  Collab. Sil& Diff. Sil\\
\textcolor{blue}{Ref: K-Means} &\textcolor{blue}{0.206 $\pm$ 0.003} &  \, \textcolor{blue}{0.000 $\pm$ 0.000} \\
\midrule
\DistrKMeans  & 0.182 $\pm$ 0.006 & -0.024 $\pm$ 0.008 \\
\Coreset  & \bfseries  0.207 $\pm$ 0.039 &\,\bfseries 0.001 $\pm$ 0.038 \\
\VPC  & 0.169 $\pm$ 0.016 & -0.037 $\pm$ 0.015 \\
\VWay  & 0.134 $\pm$ 0.044 & -0.072 $\pm$ 0.044 \\
\midrule
\Vertcohirf$^*$(Kmeans)  & 0.163 $\pm$ 0.008 & -0.043 $\pm$ 0.009 \\
\Vertcohirf$^*$(DBSCAN)  & 0.166 $\pm$ 0.009 & -0.040 $\pm$ 0.008 \\
\bottomrule
\end{tabular}
\end{sc}
\end{footnotesize}

\end{minipage}
\end{table}



\begin{table}[h]
\centering
\begin{minipage}{0.48\linewidth}
\centering

\caption{Clustering results on dataset garber-2001}
\label{tab:Garber_ARI}
\begin{footnotesize}
\begin{sc}
\begin{tabular}{lll}
\toprule
Model & Collab. ARI & Diff. ARI \\
\textcolor{blue}{Ref: K-Means} & \textcolor{blue}{0.262 $\pm$ 0.040 } &  \, \textcolor{blue}{0.000 $\pm$ 0.000} \\
\midrule
\DistrKMeans  & 0.253 $\pm$ 0.032 & -0.009 $\pm$ 0.046 \\
\Coreset  & 0.293 $\pm$ 0.058 & \, 0.031 $\pm$ 0.059 \\
\VPC  & 0.089 $\pm$ 0.024 & -0.173 $\pm$ 0.052 \\
\VWay  & 0.105 $\pm$ 0.026 & -0.157 $\pm$ 0.053 \\
\midrule
\Vertcohirf(Kmeans)  & 0.257 $\pm$ 0.025 & -0.004 $\pm$ 0.041 \\
\Vertcohirf(SC-SRGF)  &\bfseries   0.397 $\pm$ 0.034 &\,\bfseries0.136 $\pm$ 0.058 \\
\bottomrule
\end{tabular}
\end{sc}
\end{footnotesize}

\end{minipage}
\hfill
\begin{minipage}{0.48\linewidth}
\centering

\caption{Clustering results on dataset garber-2001}
\label{tab:Garber_Sil}
\begin{footnotesize}
\begin{sc}
\begin{tabular}{lll}
\toprule
Model &  Collab. Sil& Diff. Sil\\
\textcolor{blue}{Ref: K-Means} & \textcolor{blue}{0.331 $\pm$ 0.006 } &  \, \textcolor{blue}{0.000 $\pm$ 0.000} \\
\midrule
\DistrKMeans  & 0.205 $\pm$ 0.075 & -0.126 $\pm$ 0.076 \\
\Coreset  & 0.323 $\pm$ 0.009 & -0.007 $\pm$ 0.014 \\
\VPC  & 0.126 $\pm$ 0.045 & -0.205 $\pm$ 0.042 \\
\VWay  & 0.115 $\pm$ 0.048 & -0.216 $\pm$ 0.048 \\
\midrule
\Vertcohirf$^*$(Kmeans) & \bfseries 0.331 $\pm$ 0.006   &  \, \bfseries0.000 $\pm$ 0.001 \\
\Vertcohirf(SC-SRGF)  & 0.303 $\pm$ 0.067 & -0.027 $\pm$ 0.061 \\
\bottomrule
\end{tabular}
\end{sc}
\end{footnotesize}

\end{minipage}
\end{table}



\begin{table}[h]
\centering
\begin{minipage}{0.48\linewidth}
\centering

\caption{Clustering results on dataset nursery}
\label{tab:Nursery_ARI}
\begin{footnotesize}
\begin{sc}
\begin{tabular}{lll}
\toprule
Model & Collab. ARI & Diff. ARI \\
\textcolor{blue}{Ref: K-Means} & \textcolor{blue}{0.085 $\pm$ 0.012 } &  \, \textcolor{blue}{0.000 $\pm$ 0.000} \\
\midrule
\DistrKMeans  & 0.198 $\pm$ 0.101 & 0.113 $\pm$ 0.099 \\
\Coreset  & 0.171 $\pm$ 0.018 & 0.086 $\pm$ 0.026 \\
\VPC  & 0.151 $\pm$ 0.035 & 0.066 $\pm$ 0.030 \\
\VWay  & 0.128 $\pm$ 0.020 & 0.043 $\pm$ 0.013 \\
\midrule
\Vertcohirf$^*$(Kmeans)  & 0.232 $\pm$ 0.090 & 0.147 $\pm$ 0.090 \\
\Vertcohirf$^*$(KK)  & \bfseries 0.286 $\pm$ 0.114 & \bfseries 0.201 $\pm$ 0.121 \\
\bottomrule
\end{tabular}
\end{sc}
\end{footnotesize}

\end{minipage}
\hfill
\begin{minipage}{0.48\linewidth}
\centering

\caption{Clustering results on dataset nursery}
\label{tab:Nursery_Sil}
\begin{footnotesize}
\begin{sc}
\begin{tabular}{lll}
\toprule
Model &  Collab. Sil& Diff. Sil\\
\textcolor{blue}{Ref: K-Means} & \textcolor{blue}{0.077 $\pm$ 0.002 } &  \, \textcolor{blue}{0.000 $\pm$ 0.000} \\
\midrule
\DistrKMeans  & \bfseries 0.077 $\pm$ 0.002 & \bfseries -0.001 $\pm$ 0.003 \\
\Coreset  & 0.075 $\pm$ 0.003 & -0.002 $\pm$ 0.003 \\
\VPC  & 0.066 $\pm$ 0.003 & -0.011 $\pm$ 0.004 \\
\VWay  & 0.067 $\pm$ 0.002 & -0.010 $\pm$ 0.003 \\
\midrule
\Vertcohirf$^*$(Kmeans)  & 0.063 $\pm$ 0.005 & -0.014 $\pm$ 0.004 \\
\Vertcohirf$^*$(KK)  & 0.063 $\pm$ 0.008 & -0.014 $\pm$ 0.008 \\
\bottomrule
\end{tabular}
\end{sc}
\end{footnotesize}

\end{minipage}
\end{table}



\begin{table}[h]
\centering
\begin{minipage}{0.48\linewidth}
\centering

\caption{Clustering results on dataset shuttle}
\label{tab:Shuttle_ARI}
\begin{footnotesize}
\begin{sc}
\begin{tabular}{lll}
\toprule
Model & Collab. ARI & Diff. ARI \\
\textcolor{blue}{Ref: K-Means} & \textcolor{blue}{0.340 $\pm$ 0.000 } &  \, \textcolor{blue}{0.000 $\pm$ 0.000} \\
\midrule
\DistrKMeans  & 0.585 $\pm$ 0.002 & \, 0.245 $\pm$ 0.002 \\
\Coreset  & 0.469 $\pm$ 0.005 & \, 0.129 $\pm$ 0.005 \\
\VPC  & 0.398 $\pm$ 0.049 & \, 0.058 $\pm$ 0.049 \\
\VWay  & 0.327 $\pm$ 0.066 & -0.013 $\pm$ 0.066 \\
\midrule
\Vertcohirf(Kmeans)  & \bfseries  0.622 $\pm$ 0.037 & \, \bfseries 0.282 $\pm$ 0.037 \\
\Vertcohirf$^*$(DBSCAN)  & 0.545 $\pm$ 0.164 &\, 0.205 $\pm$ 0.164 \\
\bottomrule
\end{tabular}
\end{sc}
\end{footnotesize}

\end{minipage}
\hfill
\begin{minipage}{0.48\linewidth}
\centering

\caption{Clustering results on dataset shuttle}
\label{tab:Shuttle_Sil}
\begin{footnotesize}
\begin{sc}
\begin{tabular}{lll}
\toprule
Model &  Collab. Sil& Diff. Sil\\
\textcolor{blue}{Ref: K-Means} & \textcolor{blue}{ 0.680 $\pm$ 0.148 } &  \, \textcolor{blue}{0.000 $\pm$ 0.000} \\
\midrule
\DistrKMeans  & 0.598 $\pm$ 0.014 & -0.081 $\pm$ 0.135 \\
\Coreset  & 0.538 $\pm$ 0.008 & -0.141 $\pm$ 0.151 \\
\VPC  & 0.475 $\pm$ 0.121 & -0.204 $\pm$ 0.228 \\
\VWay  & 0.527 $\pm$ 0.029 & -0.153 $\pm$ 0.152 \\
\midrule
\Vertcohirf$^*$(Kmeans)  & 0.872 $\pm$ 0.136 &\,  0.193 $\pm$ 0.209 \\
\Vertcohirf(DBSCAN)  & \bfseries 0.964 $\pm$ 0.012 & \, \bfseries 0.284 $\pm$ 0.158 \\
\bottomrule
\end{tabular}
\end{sc}
\end{footnotesize}

\end{minipage}
\end{table}





\end{document}